%% file: main.tex
\documentclass[10pt,twocolumn,letterpaper]{article}

\usepackage[pagenumbers]{wacv} 
\usepackage[OT1]{fontenc} 

\makeatletter
\@namedef{ver@everyshi.sty}{}
\makeatother

\usepackage{graphicx}
\usepackage{amsmath}
\usepackage{amssymb}
\usepackage{booktabs}
\usepackage[pagebackref,breaklinks,colorlinks]{hyperref}

\usepackage{varioref}
\usepackage[export]{adjustbox}
\usepackage{amsmath}
\usepackage{bbm}
\usepackage{amssymb}
\usepackage{dsfont}
\usepackage{multirow}
\usepackage{tikz}
\usetikzlibrary{spy}
\usepackage{svg}

\usepackage[capitalize]{cleveref}
\crefname{section}{Sec.}{Secs.}
\Crefname{section}{Section}{Sections}
\Crefname{table}{Table}{Tables}
\crefname{table}{Tab.}{Tabs.}

\def\wacvPaperID{231} 
\def\confName{WACV}
\def\confYear{2024}

\begin{document}

\newcommand{\TI}{LTIM}
\newcommand{\ri}{RI}
\newcommand{\gi}{GI}
\newcommand{\gm}{GM}
\newcommand{\fft}{FFT}
\newcommand{\laion}{Laion-5B}
\newcommand{\dalT}{DALL$\cdot$E-2}
\newcommand{\dalM}{DALL$\cdot$E-Mini}

\newcommand{\RES}{Resnet-50}
\newcommand{\RESM}{Resnet-M}
\newcommand{\RESG}{Wang20}
\newcommand{\RESMG}{Grag21}
\newcommand{\RESML}{Corv22}

\newcommand{\FI}{$\mathcal{F}$}%
\newcommand{\HP}{$f_{D}$\;}%
\newcommand{\RE}{\mathbb{R}^{3 \times H \times W}}%

\newcommand{\argmin}{\mathop{\mathrm{argmin}}}

\makeatletter
\tikzset{spy on other/.code={%
  \pgfutil@g@addto@macro\tikz@lib@spy@collection{%
    \setbox\tikz@lib@spybox=\hbox{\pgfimage{#1}}}}}
\makeatother

\newcommand\blfootnote[1]{%
  \begingroup
  \renewcommand\thefootnote{}\footnote{#1}%
  \addtocounter{footnote}{-1}%
  \endgroup
}

\newcommand\niceurl[1]{%
    \href{#1}{#1}
}

\title{Deep Image Fingerprint: Towards Low Budget Synthetic Image Detection and Model Lineage Analysis}

\author{
Sergey Sinitsa, \;\;\; Ohad Fried \\
Reichman University\\
{\tt\small sergey.sinitsa@post.runi.ac.il, \;\;\; ofried@runi.ac.il}
}
\maketitle

\input{abstract.tex}
\input{intro.tex}

\input{related.tex}

\input{method.tex}

\input{experiments.tex}
\input{conclusion.tex}

{\small
\bibliographystyle{ieeetr}

\input{refer.bbl}
}

\clearpage

\appendix
\input{supp_arxiv}

\end{document}

%% file: abstract.tex
\begin{abstract}
    The generation of high-quality images has become widely accessible and is a rapidly evolving process. As a result, anyone can generate images that are indistinguishable from real ones. This leads to a wide range of applications, including malicious usage with deceptive intentions. Despite advances in detection techniques for generated images, a robust detection method still eludes us. Furthermore, model personalization techniques might affect the detection capabilities of existing methods.
    In this work, we utilize the architectural properties of convolutional neural networks (CNNs) to develop a new detection method. Our method can detect images from a known generative model and enable us to establish relationships between fine-tuned generative models. We tested the method on images produced by both Generative Adversarial Networks (GANs) and recent large text-to-image models (\TI{s}) that rely on Diffusion Models. Our approach
    outperforms others trained under identical conditions and
    achieves comparable performance to state-of-the-art pre-trained detection methods on images generated by Stable Diffusion and MidJourney, with significantly fewer required train samples.\\
    Update (June 2024): We have re-validated our method according to recent studies, which indicate that most detection methods rely only on JPEG artifacts found in real images. In this setting our results are still on par with competitors.
  
    \blfootnote{\textbf{Acknowledgement}: This work was supported in part by the Israel Science Foundation (grant No. 1574/21).}
    \blfootnote{
    \textbf{Project page}:
    \niceurl{https://sergo2020.github.io/DIF/}
    }
    
\end{abstract}

%% file: intro.tex
\section{Introduction}\label{sec:intro}

Generative neural networks enable the generation of high-quality images. While this has many benefits for scientific, creative, and business purposes, it can also be used for malicious deception. As image quality continues to improve, it becomes increasingly difficult for human observers to distinguish between real and fake images without careful observation and identification of inconsistencies, especially in spite of large text-to-image-models (\TI{s})~\cite{art:faridGeo, art:faridLight}. Therefore, there is an urgent need for an automated tool that can detect generated images. This work aims to provide such a method, which has demonstrated good performance on both novel and popular approaches.

The advancement of image generation has been made possible by the use of Deep Neural Networks (DNNs), particularly the Convolutional Neural Networks (CNNs) subclass. CNNs provide the best of both worlds by combining the image prior~\cite{art:dip} with the flexibility of DNNs~\cite{art:dnn}. This has led to the development of image generators, which are generative models trained to produce images given a sample from known distributions and can be conditioned on input. The most popular type of image generator families is the Generative Adversarial Network (GAN)~\cite{art:gan, art:dcgan}, which quickly gained popularity for its ability to produce high-quality and high-resolution images in various applications~\cite{art:biggan,art:progan,art:stylegan,art:stylegan2, art:cyclegan,art:stargan}. Previously these models have demonstrated state-of-the-art (SOTA) results in the field of image generation.

However, the advent of diffusion models~\cite{art:ddpm, art:ldm} introduced a new paradigm that has shown the ability to generate high-quality images surpassing those produced by GANs~\cite{art:ganvsddpm}. Diffusion models are also a type of generative models and can be easily conditioned on arbitrary inputs. This gave rise to a new type of image generators, the \TI{s}. These models incorporate advanced image generators (e.g.,~\cite{art:ddpm,art:vqgan, art:ldm}) and are capable of generating high-quality images from text captions combined with other input domains~\cite{art:parti, art:imagen, art:dalle2, proc:glide, art:blended, art:blendLDM, art:spa}. 
Despite the advancements and conceptual shifts, all the mentioned methods still depend on CNNs for image generation.

To detect generated images, there are typically two approaches: data-driven and rule-based. Data-driven methods~\cite{art:easycnn, art:easygan, art:universalGan} involve training models on large datasets of images, which are expected to generalize to unseen data. However, these methods may struggle to detect images from conceptually different generators, as shown in a later study~\cite{art:diffDetect}. Rule-based methods, on the other hand, rely on identifying common patterns or characteristics seen in generated images~\cite{art:ganfingerprint,art:ganfingerprintAE,art:Joslin2020, art:ganColorspaces, art:ganclues, art:graphgan, art:forencisCNN}, and typically require less data. However, most of these rules have only been demonstrated on a selected set of GAN models and image domains, which makes it necessary to re-evaluate the rules for novel types of image generators. This can be a laborious process as the rules are human-devised.

Despite significant progress in synthetic images detection, the widespread adoption of \TI{s} created new challenges in the field. Due to the resource-intensive nature of \TI{s}' training, access to \TI{s} and the amount of available generated images for research purposes is limited. As a result, data-driven approaches and some rule-based methods may not be practical, as they require a significant amount of data for training, and in some cases the training of an image generator itself. Therefore, new detectors are expected to not only perform well, but also to be trained on a small number of images.

Detecting images generated by ``personalized'' models poses an additional challenge. These models are fine-tuned versions of \TI{s} that are trained to generate images with specific objects or in particular styles~\cite{art:dreambooth,art:textual, art:break}. Despite their widespread use, the impact of personalization and fine-tuning on model fingerprints has not been studied.

The proposed method achieves high detection accuracy with minimal training data (less than 512 images), leveraging a CNN architecture to extract fingerprints of image generators and detect images from the same generator. Other detectors typically require hundreds of thousands of images.

Our method was tested on various image generators, including established GANs and \TI{s}. It outperforms competitors trained under the same conditions and achieves comparable performance to SOTA pre-trained detectors on widely used \TI{s} such as Stable Diffusion~\cite{art:ldm} and MidJourney~\cite{misc:midJ}, with significantly fewer required training samples. Moreover, our method proves valuable for model lineage analysis (\cref{sec:lineage}). However, it is important to note that our method is a proof-of-concept work and has certain limitations, which are discussed in~ \cref{sec:conclusion}.

Recently, a new study~\cite{art:jpeg} revealed that current detection methods are heavily biased on JPEG artifacts. Following this development, we further evaluate with unbiased JPEG data, and show solid performance on par with other methods.

%% file: related.tex
\section{Related Work}\label{sec:related}

Data-driven methods rely on CNN classification models. These methods aim to detect compressed images from unknown image generators by training a detector on images from a single image generator. The authors of \cite{art:easycnn} fine-tune a pre-trained \RES~\cite{art:resnet} with 720k fake and real images produced by ProGAN and apply a set of compressions during training. They achieve high average precision on images from several GAN models, but demonstrate poor accuracy (\cref{sec:detection}). In \cite{art:easygan}, the authors repeat the process, but with a modified Resnet50 and heavier augmentations, resulting in SOTA performance. However, after the development of \TI{s}, another study~\cite{art:diffDetect} revealed that later model generalizes well only on the same image generator family.

One notable approach from rule-based methods involves detecting spatially-stationary and high-frequency artifacts that image generators produce within images~\cite{art:checkerboard}. These artifacts were observed in generated images~\cite{art:ganfingerprint,art:ganfingerprintAE,art:Joslin2020}, including those produced by \TI{s}~\cite{art:diffDetect, art:intrigue}. Since they are unique to each trained image generator, they are referred to as \emph{fingerprints} (\FI{-s}). To estimate \FI, a set of residuals is produced by passing each image through a denoising filter (\HP), and the residuals are then averaged. This leaves only the common deterministic pattern within the residuals --- fingerprint. The image is associated with the generative model by calculating the correlation coefficient between its residual and the model's \FI.

The usage of \FI{-s} holds great potential in model lineage analysis. Marra~et~al.~\cite{art:ganfingerprint} demonstrated that residuals of images generated by a specific model architecture exhibit high correlation not only with the model's fingerprint but also with the fingerprints of models sharing the same architecture \cite{art:ganfingerprint}. 
Yu~et~al. proposed a supervised method to attribute fake images to their source models, training a detector on a dataset of images generated by multiple GANs \cite{art:ganfingerprintAE}. Nevertheless, this approach relies on manual supervision, requiring researchers to make educated guesses about the relationship between different models prior to training. Previous studies have not explored 
the relationship between models and their fine-tuned versions, or the methodology for model lineage analysis.

In addition to the above, \FI{-s} and other artifacts are primarily observed in the spectrum space, by transforming the image with the Fast Fourier Transform~(\fft). Some studies~\cite{art:spec, art:ganfreq} use this concept by training models on image spectrum samples or performing operations within Fourier space. Some attempts have also been made to synthesize a large set of \FI{-s} for further training of detectors~\cite{art:spec, inp:fingerprintNet}. However, while these methods demonstrate good detection accuracy for a set of image generators on average, they may perform poorly with some of them.

Another group of rule-based methods focuses on detecting color distortions in generated images~\cite{art:ganColorspaces, art:ganclues, art:graphgan, art:forencisCNN}. These detection methods are usually evaluated on both generated and natural images and have shown promising results. However, it should be noted that color distortions have only been demonstrated for a few image domains and GAN models.

%% file: method.tex
\section{Method}\label{sec:method}

Our method extracts a CNN fingerprint of an image generator using a small number of generated images and applies it for the detection of other images produced by the same image generator. The method relies on the properties of CNNs, as explained below with a simple experiment~(\cref{sec:induce}). Then we'll explain the fingerprint extraction~(\cref{sec:train}) and following detection process~(\cref{sec:inference}), including implementation details~(\cref{sec:implement}).

\subsection{Deep Image Fingerprint}\label{sec:induce}

Deep Image Prior~\cite{art:dip} demonstrated that CNNs incorporate an image prior in their structure. Consider an image restoration task, where given a corrupt image $X \in \RE$ the goal is to obtain the restored image $\hat{Y} \in \RE$ from a trained model $h_{\theta}$:
\begin{equation}
   \hat{Y} = \min_{\theta} E(h_{\theta},X ) + P(X)
   \label{eq:genImage}
\end{equation}
$E$ is a data term and $P$ is the image prior.

For a CNN encoder-decoder model $g_{\theta}$, the architecture itself serves as the image prior.
Thus, the optimization task is simplified, as the search is focused on finding only the data term $E$ within the weights ($\theta$) space of the CNN without the need of learning $P$. Here the input of $g_{\theta}$ is a random tensor $Z \in \mathbb{R}^{C \times H \times W}$, where each element is $z_i \sim U(0,1)$.
\begin{equation}
   \hat{Y} = \min \limits_{\theta} E(g_{\theta}(Z), X)
   \label{eq:dipOpt}
\end{equation}

After the optimization, the reconstructed image is given by: 
\begin{equation}
   \hat{Y} = g_{\hat{\theta}}(Z), \;\;\; \hat{\theta} = \argmin_{\theta} E(g_{\theta}(Z), X)
   \label{eq:dipOutput}
\end{equation} 
But, following a number of observations, 
\cref{eq:dipOutput} seems to be incomplete. Images produced by CNNs include a unique model fingerprint (\FI$\in \RE$), thus: 
\begin{equation}
   \hat{Y} + \mathcal{F}(g_{\hat{\theta}}) = g_{\hat{\theta}}(Z)
\end{equation}
To prove this statement, we perform a simple experiment: we optimize the weights of CNN ($\theta$) U-Net~\cite{art:unet} to produce a single gray image without any semantic content. It seems as a trivial task, yet after convergence, the model is still unable to reconstruct the image perfectly (\cref{fig:monochromeExp}). 
\input{figs/figures/blank_v2}

We reveal the artifacts by simple image normalization to range [0,1]. Two main fingerprint patterns are observed: up-sampling and boundary artifacts. Previous research has primarily focused on up-sampling artifacts, resulting from interpolation and kernel overlap \cite{art:checkerboard, art:ganfreqfix, art:ganfreq, inp:fingerprintNet}. In general, an up-sampling replicates signal in spectrum domain. To simulate only these artifacts, we optimize the Up-Net model: four blocks of 1x1~\cite{art:OxO} convolutional layers, to avoid padding, followed by deconvolutional layers with kernel size of 2 and stride 2. This produces a periodic pattern in image space and dots in the spectrum domain (\cref{fig:monochromeExp}).

Boundary artifacts cause a grid-like structure in spectrum domain. When applying \fft{} to an image, it assumes periodicity, but if image is not periodic, a ``cross'' artifact will appear as a result of the image's non-periodicity~\cite{art:priodicplus}. Our target gray image, which is periodic, does not exhibit this artifact, but the reconstruction does. This phenomenon is rooted in image padding and the mechanism of convolution layers, which are known to impact CNN performance \cite{art:mindpad,art:learningpad}. 
To simulate this, we optimize the C-Net model with eight convolutional layers, each having a kernel size of 3, a stride of 1, and a padding of 1. This configuration helps preserve the spatial dimensions of the input and induces artifacts.
Addition of up-sampling replicates ``cross'' structure resulting in grid structure in spectrum domain.

Consequently, in parallel to the Deep Image Prior~\cite{art:dip}, where a CNN's structure was shown to be an image prior, here we have shown that CNN's structure is also artifact prior. As such, we term our method \emph{deep image fingerprint} (DIF), and describe it next.

\subsection{Fingerprint Extraction}\label{sec:train}

We can extract fingerprints of a target model by an optimization of $\theta$ given a set of generated images and a set of arbitrary real images. The optimization is similar to the denoising procedure~\cite{art:dip}, where \FI{} is acquired by \cref{eq:dipOutput}, but instead of computing mean square error loss with respect to some image we compute correlation with respect to a set of image residuals. Residual $R_i \in \RE$ is defined as $R_i = f_{D}(X_i)$, where ($X_i$) is an image and \HP is a denoiser filter~\cite{art:ganfingerprint}.

The goal is to produce fingerprint that is highly correlated with residuals of generated images, and non-correlated with residuals of real images. Pearson Correlation Coefficient is proved to be a good correlation metric between image residual and model's fingerprint~\cite{art:ganfingerprint, art:ganfingerprintAE}. In practice we transform each input into their zero-mean and unit-norm versions, preform inner product and average values. This will be referred simply as \emph{correlation} and denoted as $\rho(\cdot, \cdot)$.

The loss function is formulated similarly to contrastive loss in siamese setting~\cite{art:contrastive}. We define a similarity factor $t_{ij} \in \{0,1\}$, which is equal to 1 when two residuals are from the same class and 0 otherwise and correlation distance ($D_{ij}$) as euclidean distance between correlation values:
\begin{equation}
   D_{ij} = \sqrt{(\rho(R_i, \mathcal{F}) - \rho(R_j, \mathcal{F}))^2}
   \label{eq:corrDist}
\end{equation}
Finally the sample loss function ($\mathcal{L_{S}}$) is summarised below:
\begin{equation}
   \mathcal{L_{S}} =\frac{ t_{ij} \cdot D_{ij} + (1 - t_{ij}) \cdot \left( m - D_{ij}
    \right)}{m}
    \label{eq:lossS}
\end{equation}
$m$ is a hyper parameter. 

\subsection{Detection of Generated Images}\label{sec:inference}

We perform detection through hypothesis testing. 
After generating the \FI{} from the trained model $(g_{\hat{\theta}})$, the model itself (and a GPU) is no longer necessary. 
Then, we compute the means of the populations of real and fake correlations according to \cref{eq:refVals}, denoted as $\mu_{\boldsymbol{r}}$ and $\mu_{\boldsymbol{g}}$, respectively. $N_{\boldsymbol{r}}$ and $N_{\boldsymbol{g}}$ represent the number of real and generated images in the training set. For each test image, we extract its residual $R_{\text{test}} = f_{D}(X_{\text{test}})$ and then perform hypothesis testing as follows: if $|\rho( R_{test}, \mathcal{F} ) - \mu_{\boldsymbol{g}}| < |\rho(R_{test}, \mathcal{F}) - \mu_{\boldsymbol{r}}|$, the tested image is considered generated, otherwise, it is considered real or not produced by the target model. This procedure does not require any parameters, such as threshold.
\begin{equation}
   \mu_{\boldsymbol{r}} = \frac{1}{N_{\boldsymbol{r}}} 
   \sum_{i \in \boldsymbol{r}} \rho(R_{i}, \mathcal{F})
   \;,\;\;
   \mu_{\boldsymbol{g}} = \frac{1}{N_{\boldsymbol{r}}} 
   \sum_{i \in \boldsymbol{g}} \rho(R_{i}, \mathcal{F})
    \label{eq:refVals}
\end{equation}

\subsection{Implementation Details}\label{sec:implement}

In all of our experiments (\cref{sec:experiments}), we train and test the method in a similar manner. Margin is constant $m=0.01$ and $z$ has 16 channels. \HP is a pre-trained DnCNN model~\cite{art:dncnn} that is trained separately on real images (\laion{} dataset \cite{misc:laion5b}) with sigma range of $[5,15]$ and crop size $48 \times 48$ pixels. During the extraction procedure, \HP{'s} weights are not updated. The extraction model is a U-net~\cite{art:unet}. Optimization was carried out using Adam~\cite{art:adam}, with a constant learning rate of $5e^{-4}$. During training, the fingerprint was accumulated using exponential moving averaging. We provide additional details about the selection of \HP{}, the training process of DnCNN, and the U-Net architecture in our supplementary materials.

%% file: figs/figures/blank_v2.tex
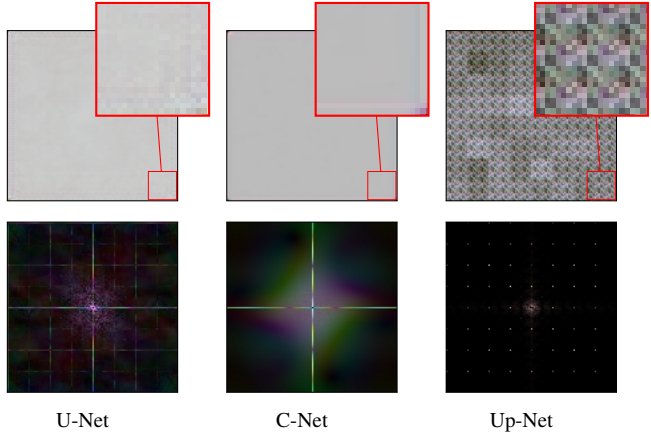
\begin{figure}[t]
\captionsetup[subfigure]{labelformat=empty}

    \begin{subfigure}{0.3\linewidth}
    \centering
    \begin{minipage}{0.9\textwidth}
        \begin{tikzpicture}[spy using outlines={rectangle,red,magnification=4,size=1.5cm, connect spies}]
        \node {\includegraphics[width=1\textwidth,frame]{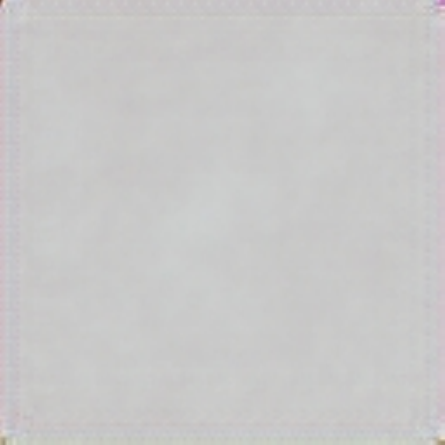}};
        \tikzset{spy on other={
        \node {\pgfimage[width=1\textwidth]{figs/imgs/conv/2000_unet.pdf}};}}
        \spy on (0.93,-0.93) in node [below] at (0.8,1.5);
        \end{tikzpicture}
        \begin{tikzpicture}[auto]
        \node {\includegraphics[width=1\textwidth,frame]{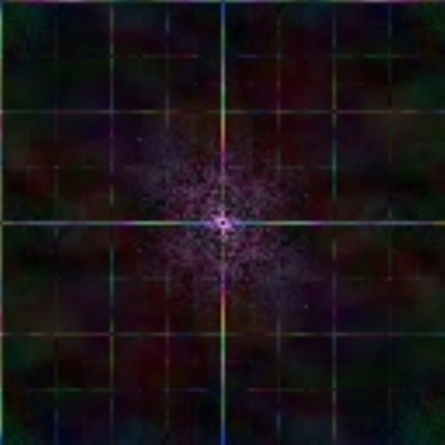}};
        \end{tikzpicture}
    \end{minipage}
    \caption{U-Net}
    \end{subfigure}%
    \hfill
    \begin{subfigure}{0.3\linewidth}
    \centering
    \begin{minipage}{0.9\textwidth}
        \begin{tikzpicture}[spy using outlines={rectangle,red,magnification=4,size=1.5cm, connect spies}]
        \node {\includegraphics[width=1\textwidth,frame]{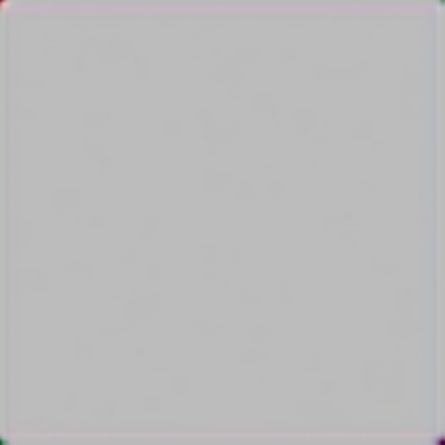}};
        \tikzset{spy on other={
        \node {\pgfimage[width=1\textwidth]{figs/imgs/conv/2000_conv.pdf}};}}
        \spy on (0.93,-0.93) in node [below] at (0.8,1.5);
        \end{tikzpicture}
        \begin{tikzpicture}[auto]
        \node {\includegraphics[width=1\textwidth,frame]{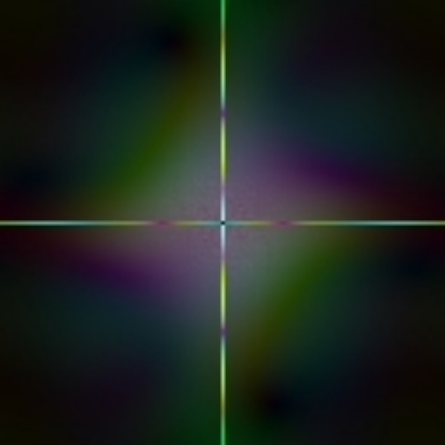}};
        \end{tikzpicture}
    \end{minipage}
    \caption{C-Net}
    \end{subfigure}%
    \hfill
    \begin{subfigure}{0.3\linewidth}
    \centering
    \begin{minipage}{0.9\textwidth}
        \begin{tikzpicture}[spy using outlines={rectangle,red,magnification=4,size=1.5cm, connect spies}]
        \node {\includegraphics[width=1\textwidth,frame]{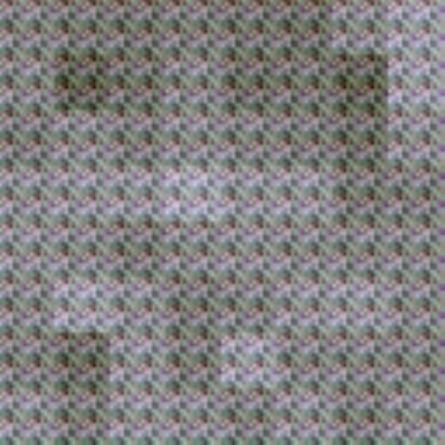}};
        \tikzset{spy on other={
        \node {\pgfimage[width=1\textwidth]{figs/imgs/conv/2000_up.pdf}};}}
        \spy on (0.93,-0.93) in node [below] at (0.8,1.5);
        \end{tikzpicture}
        \begin{tikzpicture}[auto]
        \node {\includegraphics[width=1\textwidth,frame]{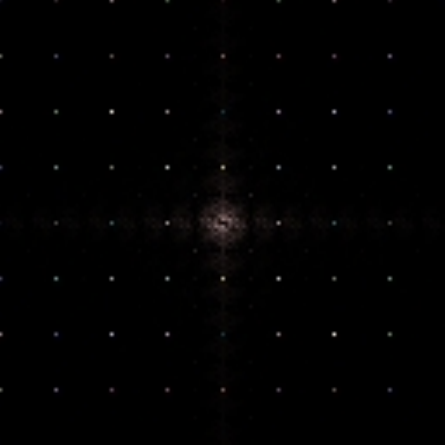}};
        \end{tikzpicture}
    \end{minipage}
    \caption{Up-Net}
    \end{subfigure}%
     \caption{
     The reconstructed gray images (top) and their \fft{} log-magnitude (bottom). For previewing images are normalized. The mean value is subtracted before applying the \fft{}. U-Net produces mix of boundary artifacts (lines) and up-sampling artifacts (checkerboard). C-Net produces solely boundary artifacts, while Up-Net exclusively yields up-sampling artifacts. In the Up-Net model, the input noise is up-sampled after 1x1 convolution, resulting in blocks with varying gray levels.}
     \label{fig:monochromeExp}
\end{figure}%

%% file: experiments.tex
\section{Experiments}\label{sec:experiments}

This study includes a series of experiments that involve a varied collection of images generated by different \TI{s} and GANs. The datasets are summarized in \cref{sec:data}. The detection results are presented in \cref{sec:detection}, which is divided into two parts: the detection of images produced by \TI{s} and the detection of GAN-generated images. In \cref{sec:source}, we investigate the effect of image generator training and fine-tuning on its fingerprint, and in \cref{sec:lineage}, we analyze the relationship between selected \TI{s}. Finally, in \cref{sec:robustness}, we test the method on compressed images.

\subsection{Detection Data}\label{sec:data}

Our data includes generated images from a variety of \TI{} and GAN models. In contrast to \TI{s}, that produce multi-domain images, GANs are often limited to a specific image domain that they were trained on. Therefore, some of the GAN datasets include a number of image domains, each produced by a different model. \cref{tab:data} summarizes the data. The amount of real and generated images per set is equal. During the experiments we randomly split each dataset into equally sized train and test sets.
\input{figs/tables/data.tex}

Datasets representing \TI{s} were generated by us and is available online. We randomly selected real images and their corresponding captions from the \laion~\cite{misc:laion5b} dataset. Then, we used the captions to generate corresponding images using publicly available Stable Diffusion models~\cite{art:ldm}, versions 1.4 (SD 1.4) and 2.1 (SD 2.1), \dalM~\cite{misc:dallemini} and GLIDE~\cite{proc:glide}. In the case of \dalT~\cite{art:dalle2} we generated images with OpenAI's official API\footnote{
\niceurl{https://platform.openai.com/docs/guides/images}}. Lastly, a large set of generated images produced by MidJourney (MJ)~\cite{misc:midJ} is publicly available on the Kaggle website~\cite{misc:midJdata}, from which we select a random subset for this work. 

Datasets representing GAN models were obtained from the supplementary materials of Wang~et~al.~\cite{art:easycnn}. Our supplementary document contains a detailed review of the datasets.

\subsection{Detection of Generated Images}\label{sec:detection}

We evaluate DIF as a generated image detector and compare it to rule-based methods and data-driven methods. We summarize the results of detecting images produced by \TI{s} and GANs in \cref{tab:accuracyTI,tab:accuracyGAN}, respectively. To represent the number of training samples and the size of the pre-training datasets, we utilize $N_S$ and $N_D$ respectively.
\input{figs/tables/comparisonTI_v2}

Rule-based methods rely on fingerprint extraction. Marra18~\cite{art:ganfingerprint} is a conventional method for extracting \FI{} that involves averaging over residuals. Classification is then performed using the extracted \FI{} as described in \cref{sec:detection}. Joslin20~\cite{art:Joslin2020} follows a similar approach to Marra18, but correlation is conducted in the Fourier domain. For these two methods and DIF, we employ the same \HP{}~(\cref{sec:implement}), as with it they demonstrate superior performance. Further details can be found in the supplementary material. Ning18~\cite{art:ganfingerprintAE} involves extracting fingerprints from images using a trainable auto-encoder model and classifying the images with an additional CNN model.

We also compare to data-driven methods: pre-trained \RES~\cite{art:resnet} on ImageNet~\cite{art:imagenet}, \RESG~\cite{art:easycnn} is \RES, which is trained on ProGAN images, \RESMG~\cite{art:easygan} and \RESML~\cite{art:diffDetect} utilize modified \RES{} architecture (\RESM{}).

The methods \RESG{}, \RESMG{}, and \RESML{} have previously demonstrated SOTA results in the detection of images produced by GANs and LDMs~\cite{art:diffDetect}.  \RESG{}, \RESMG{}, and \RESML{} are trained on 720k images each, where half of the dataset consists of real images, while the other half is composed of images generated by ProGAN~\cite{art:progan}, StyleGAN~\cite{art:stylegan} and LDM~\cite{art:ldm}, respectively. Additionally, we performed fine-tuning on all the aforementioned models using $N_S$ images. In the case of fine-tuning, the models are frozen, and only the last fully-connected layer is reinitialized and updated during training.

According to the results in \cref{tab:accuracyTI}, the proposed method achieves remarkably high performance with a low number of training samples when applied to images generated by \TI{s}. DIF demonstrates higher accuracy compared to other rule-based methods and a higher mean accuracy compared to non pre-trained data-driven methods. Furthermore, when directly comparing DIF to the best fine-tuned method (\RESML{}), DIF performs equally well in the case of SD 1.4, MJ, and \dalM{}, and outperforms all non fine-tuned models on average, even with $N_S = 128$.

For GLIDE, SD 2.1 and \dalT{}, the detection accuracy of DIF is lower. \FI{} is formed by model architecture, weight initialization, and train data~(\cref{sec:related}). Refer to \cref{fig:examplesExt}. In contrast to SD 1.4, the $\mathcal{F}_A$ of GLIDE exhibits a barely visible grid pattern that gets lost within the accumulated noise. This makes it a ``weak'' fingerprint, characterized by lower energy. Similar ``weak'' fingerprint is also observed in SD 2.1. Despite sharing architecture with SD 1.4, SD 2.1 was trained on different dataset\footnote{\label{foot:sd} \niceurl{https://github.com/Stability-AI/stablediffusion}} and possibly with different weight initialization.
\input{figs/figures/ffts}

Similar results are obtained when evaluating images generated by GANs, as shown in \cref{tab:accuracyGAN}. DIF surpasses all rule-based methods and performs comparably to the best pre-trained model (\RESMG{}) in four out of seven GAN models.
The lowest detection accuracy is observed for the ProGAN$_e$ dataset, which is expected. As mentioned in \cref{sec:related}, each trained model produces unique fingerprints, and DIF is specifically designed to detect images from a single model. For completeness, we also measure the detection accuracy for ProGAN$_t$, where DIF is trained on images of each model separately. In this case, the mean accuracy is 91.2\%, with the highest accuracy reaching 96.0\% and the lowest accuracy being 83.2\%. Similarly to the case of SD models, we observe that not only model architecture affects it's \FI{}. A comprehensive evaluation for ProGAN$_t$ is provided in our supplementary materials. 

Summarizing the experiments, DIF exhibits strong performance with both novel and well-established image generation methods. We have demonstrated that DIF outperforms other methods trained under the same conditions on average, in the detection of images generated by both \TI{}s and GANs. Additionally, DIF performs comparably to pre-trained SOTA methods (\RESMG{} and \RESML{}). It is worth noting that DIF achieves these results with just a few hundred images, while pre-training typically requires three orders of magnitude more images and additional fine-tuning. Therefore, we consider DIF to be a formidable competitor to existing methods.
\input{figs/tables/comparisonGAN}
\subsection{Detection of Images From Fine-Tuned Models}\label{sec:source}

Now consider a more challenging setting, where given a trained DIF for some image generator we aim to detect images generated by its fine-tuned version. Due to uniqueness of fingerprints (\cref{sec:related}) images generated by these new model variations might not be detected by DIF trained on images produced by the original model.

To investigate the relation between source models and their variations we conduct an experiment using varied datasets and checkpoints. First, we train four ProGAN models ($P_{A}, P_{B}, \hat{P}_{A}$, and $\hat{P}_{B}$) for 70 epochs. $P_{A}$ and $P_{B}$ are trained on 2,500 ``cat'' class images from AFHQ~\cite{art:stylegan2} with random seeds A and B respectively. $\hat{P}_{A}$ and $\hat{P}_{B}$ are trained with the same random seeds, but on 2,500 ``wild'' class images from AFHQ. Next, for each model we use five checkpoints at epochs 20, 32, 40, 52 and 70, and generate 2,500 images for each. Finally, we construct 20 datasets by adding real images from the corresponding train set of ProGANs to each set of generated images. This results in 20 ProGAN models with 20 datasets.
\input{figs/figures/cats_cross_eval}
We preform a cross-detection on $P_{A}, P_{B}, \hat{P}_{A}$ and $\hat{P}_{B}$ and for brevity summarize results for $P_{A}$ and $P_{B}$ in \cref{fig:crossevalCATS}. The full cross-detection map, including comparison to cross-correlation of fingerprints, is available in our supplementary materials. During cross-detection we attempt to classify images produced by some image generator with DIF which is trained on images from another image generator. Observe the symmetric relation within the same model: for checkpoints of epochs 40, 52 and 70 cross-detection accuracy is high and symmetric. Other relations include low accuracy and/or asymmetric, which is exactly what is expected for DIF trained on unique \FI{-s}. We may conclude that the model's \FI{} changes during training. However, as the model converges, these changes become insignificant, resulting in high cross-detection accuracy for DIF.

We conduct an additional experiment where we measure cross-detection on images from a number of fine-tuned/``personalized'' stable diffusion models~\cite{art:dreambooth}, which involve updates of image decoder weights. The models are: our custom fine-tuned SD 1.4 with a small set of images and two downloaded models of stable diffusion v1.5 and v2.0 fine-tuned on a large set of anime images\footnote{\niceurl{https://huggingface.co/DGSpitzer/Cyberpunk-Anime-Diffusion}} and robot images\footnote{\niceurl{https://huggingface.co/nousr/robo-diffusion-2-base}}. Models denoted as SD 1.4S, SD 1.5A and SD 2.0R respectively. For each model we generated 1000 images from the same caption set that was used with SD 1.4, including style/object keywords specific to each model. Then train DIF for each with 512 real and 512 generated images (\cref{sec:detection}). Upon analyzing the results presented in \cref{fig:crossevalSD}, we can conclude that the relationships observed in previous experiments are maintained even when the model is fine-tuned with new data. Consequently DIF will also perform well on images generated by fine-tuned models.
\input{figs/figures/cross_eval_sd}
\subsection{Model Lineage Analysis}\label{sec:lineage} 

We can use our extracted fingerprints to detect the lineage of trained models. In \cref{fig:crossevalTI} we show cross-detection results for \TI{s}. We observe that SD 1.4 and MJ produce high-cross detection accuracy, thus MJ is likely to be a fine-tuned version of SD 1.4. Indeed, while this is not public knowledge, we found evidence that our analysis is correct\footnote{\niceurl{https://tokenizedhq.com/midjourney-model/}}.

In contrast, SD 2.1 does not retain such a relation with both SD 1.4 and MJ, therefore we can conclude that this model was trained from scratch. Indeed, this was confirmed by the SD 2.1 developers\footnotemark[3].
\input{figs/figures/cross_eval_t2i}

\subsection{Robustness}\label{sec:robustness}
 
To test our method's robustness to image compression, we use two models: SD 1.4 and GLIDE, representing strong and weak \FI{-s}, respectively. We created four additional datasets for each model: images compressed using JPEG at quality levels of 75 (J75) and 50 (J50), resized (R) images, and blurred (B) images. Uncompressed images are denoted as U. The resized images were down-sampled to half of their original size and then up-sampled back to their original size using nearest-neighbor interpolation. We applied blur with a sigma value of 3, resulting in heavy smoothing.

We train DIF separately on each compression set for each model, following the procedure outlined in \cref{sec:detection}, using 256 training images each time. To emulate a more realistic scenario in which the compression type is not known, we also train the model on a mixed dataset where images are randomly compressed during the training procedure.
\input{figs/figures/compression}

The detection accuracy for each compression type in reported in \cref{fig:compressSD14,fig:compressGLIDE} for SD 1.4 and GLIDE, respectively. We observe that the blur significantly reduces the extraction and detection capabilities of the method because the \FI{} signal resides on higher frequencies of the image. In other cases, the accuracy is reduced, but this depends on the characteristics of the fingerprint. 

We hypothesize that the detection of compressed images is influenced by the original pattern of \FI{}. For example, we can observe different detection behaviors for resized images in both models. DIF trained on GLIDE's resized images can easily detect uncompressed images, but in the same scenario, where DIF is trained on images produced by SD 1.4, it is unable to detect uncompressed images. Additionally, we can barely observe symmetry in cross-detection accuracy per model. This is a clear sign of changes within \FI{} that result from compression.

The above is only a brief analysis of the effect of compression on \FI{}. It appears to be a complex topic that requires a more comprehensive investigation, which we leave for future work.

As suggested by a recent study~\cite{art:jpeg}, we compare our method to others in a more challenging setting, where generated images are also compressed. We observe that the mean and median JPEG quality of the \laion{} dataset are 0.84 and 0.85, respectively. Therefore, we compress the generated images using the median quality factor of 0.85 and perform experiments from \cref{sec:detection} for DIF, \RESM{}, and \RESML{} with 1024 samples. From~\cref{tab:jpegTI}, we conclude that despite performance degradation, DIF remains competitive.
\RESML{} dominates in this comparison due to the compression applied as part of its training augmentations. However, it requires a large pre-training dataset.

\input{figs/tables/jpeg_ti}

%% file: figs/tables/data.tex

\begin{table}[ht]
    \centering
    \footnotesize
    \begin{tabular}{ lcc }
     \toprule 
     \textbf{Source model} & $\boldsymbol{N_I}$ & $\boldsymbol{N_M}$\\
     \midrule
     CycleGAN & 2,600 & 6 \\
     
     ProGAN$_e$ & 8,000 & 20 \\

     ProGAN$_t$ & 80,000 & 20 \\
     
     BigGAN & 4,000 & 1 \\
     
     StyleGAN & 12,000 & 3 \\
     
     StyleGAN2 & 16,000 & 4 \\
     
     GauGAN & 10,000 & 1 \\
     
     StarGAN & 4,000 & 1 \\
     
     SD 1.4 & 6,000 & 1 \\
     
     SD 2.1 & 6,000 & 1 \\
     
     MJ & 6,000 & 1 \\

     \dalT & 2,000 & 1 \\

     GLIDE & 6,000 & 1 \\

     \dalM & 6,000 & 1 \\
     
     \bottomrule
    \end{tabular}
    \caption{Our data. 
    We specify the source model, number of images $N_I$, and number of model variants $N_M$.
    \label{tab:data}}
\end{table}

%% file: figs/tables/comparisonTI_v2.tex

\begin{table*}[ht]
    \centering
    \footnotesize
    \begin{tabular}{ cclccccccc }
    \toprule 
    $\boldsymbol{N_P}$ & $\boldsymbol{N_S}$ & \textbf{Method} & \textbf{SD 1.4} & \textbf{SD 2.1} & \textbf{MJ} &\textbf{\dalM}& \textbf{GLIDE} & \textbf{\dalT} & \textbf{Mean}\\
    \midrule
    \multirow{14}{*}{\textbf{0}}  &  \multirow{6}{*}{1024} 
    & Joslin20 & 49.7 & 49.9  & 49.9 & 52.3 & 57.0 & 51.4 & 51.7 \\
    & & Marra18 & 52.6 & 48.0  & 75.7 & 85.3 & 57.6 & 56.3 & 62.6  \\
    & & Ning18 & 50.8 & 51.4  & 59.5 & 58.1 & 57.7 & 52.2 & 55.0 \\
    & & \RES & 72.2 & 72.6 & 87.1 & 87.5 & 94.1 & \textbf{88.7} & 83.7  \\
    & & \RESM &  69.2 & 73.6 & 89.7 & 89.4 & \textbf{94.3} & 85.9  & 83.7 \\
    & & DIF & \textbf{99.3} & \textbf{89.5}  & \textbf{99.0} & \textbf{99.0} & 90.3 & 79.5  & \textbf{92.8} \\
    \cmidrule{2-10}
    
    & \multirow{3}{2.em}{512} 
    & \RES & 70.6 & 70.5 & 90.0 & 85.4 & \textbf{92.7} & \textbf{85.4}  & 82.4 \\
    & & \RESM &  68.3 & 71.4 & 81.1 & 85.2 & 91.3 & 83.8 & 80.2  \\
    & & DIF & \textbf{99.2} & \textbf{86.3}  & \textbf{98.8} & \textbf{98.7} & 88.2 & 79.1  & \textbf{91.7} \\
    \cmidrule{2-10}
    
    & \multirow{3}{2.em}{256} 
    & \RES & 66.2 & 64.0 & 86.4 & 82.0 & \textbf{89.0} & \textbf{81.4}  & 78.2  \\
    & & \RESM &  67.2 & 65.8 & 74.4 &  83.5 & 88.0 & 75.3  & 75.7 \\
    & & DIF & \textbf{98.5} & \textbf{81.3}  & \textbf{98.1} & \textbf{98.0} & 85.9 & 77.7  & \textbf{89.9} \\
    \cmidrule{2-10}

    & \multirow{3}{2.em}{128} 
    & \RES & 66.2 & 64.0 & 85.9 & 75.1 & \textbf{87.6} & 75.2  & 75.7 \\
    & & \RESM &  68.0 & 66.4 & 72.6 & 81.0 & 80.3 & 71.7  & 73.3\\
    & & DIF & \textbf{97.7} & \textbf{75.5}  & \textbf{97.3} & \textbf{97.0} & 81.4 & \textbf{76.1}  & \textbf{87.5} \\
    \midrule

    \multirow{11}{*}{\textbf{720k}}  &  \multirow{3}{*}{1024} 
    & \RESG &  63.7 & 61.7 & 75.7 & 78.3 & 74.5 & 74.7 & 71.4  \\
    & & \RESMG & 93.5 & 86.9 & 93.5 & 96.1 & \textbf{97.5} & \textbf{93.5}  & 93.5 \\
    & & \RESML &  \textbf{99.7} & \textbf{99.0} & \textbf{99.2} & \textbf{96.4} & 96.0 & 91.9  & \textbf{97.0} \\
    \cmidrule{2-10}
    
    &  \multirow{2}{*}{512} 
    & \RESMG  &  93.0 & 86.0 & 94.7 & \textbf{96.1} & \textbf{96.4} & \textbf{92.1}  & 93.1  \\
    & & \RESML &  \textbf{99.6} & \textbf{98.6} & \textbf{98.8} & 95.9 & 95.1 & 89.4  &\textbf{96.2} \\
    \cmidrule{2-10}

    & \multirow{2}{*}{256} 
    & \RESMG  &  89.9 & 82.9 & 94.1 & 94.9 & \textbf{95.6} & \textbf{90.5}  & 91.3 \\
    & & \RESML &  \textbf{99.6} & \textbf{98.6} & \textbf{99.0} & \textbf{95.6} & 94.8 & 89.0  & \textbf{96.1} \\
    \cmidrule{2-10}

    & \multirow{2}{*}{128}
    & \RESMG  &  87.9 & 82.2 & 93.4 & 93.5 & \textbf{94.2} & \textbf{90.4}  & 90.3 \\
    & & \RESML &  \textbf{99.6} & \textbf{98.4} & \textbf{98.6} & \textbf{93.9} & 89.1 & 77.5 & \textbf{92.9} \\
    \cmidrule{2-10}

    & \multirow{2}{*}{0}
    & \RESMG &  57.0 & 50.2 & 63.1 & 58.0 & 54.3 & 51.2  & 55.6 \\
    & & \RESML &  \textbf{99.3} & \textbf{99.3} & \textbf{99.0} & \textbf{89.5} & \textbf{57.0} & \textbf{51.6}  & \textbf{82.6} \\
    
     \bottomrule
    \end{tabular}
    \caption{Classification accuracy (\%). $N_S$ and $N_P$ are the amount of train samples and pre-train dataset size. DIF achieves best average accuracy comparing to other rule-based and non pre-trained models even when $N_S = 128$. For SD 1.4, MJ and \dalM{}, DIF is also comparable with the SOTA pre-trained and fine-tuned method (\RESML{}).}
    \label{tab:accuracyTI}
\end{table*}

%% file: figs/figures/ffts.tex
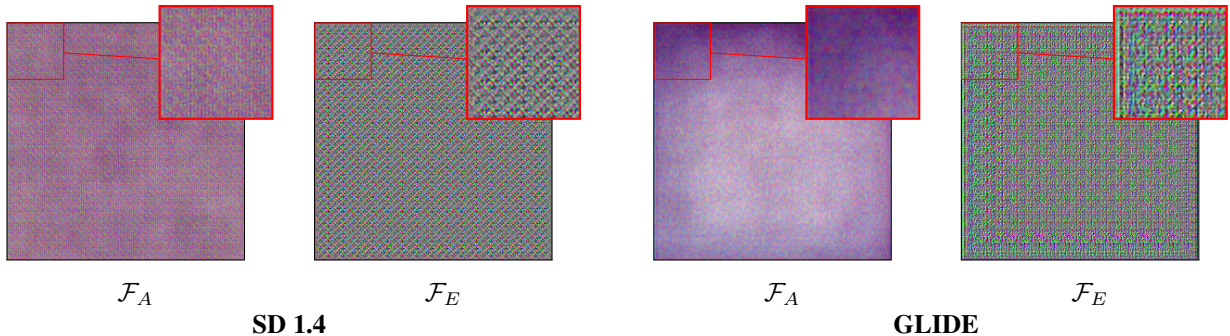
\begin{figure*}[t]
\centering
\begin{tabular}{ccccc}
    \centering

    \begin{tikzpicture}[spy using outlines={rectangle,red,magnification=2.,size=1.5cm, connect spies}]
    \node {\includegraphics[width=0.18\textwidth,frame]{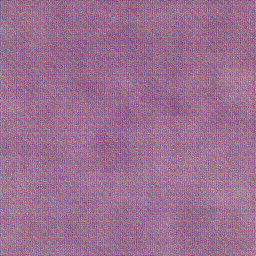}};
    \tikzset{spy on other={
    \node {\pgfimage[width=0.18\textwidth]{figs/imgs/mean_models/finger_sd14_fake.png}};}}
    \spy on (-1.2,1.2) in node [below] at (1.2,1.8);
    \end{tikzpicture}
    & 

    \begin{tikzpicture}[spy using outlines={rectangle,red,magnification=2.,size=1.5cm, connect spies}]
    \node {\includegraphics[width=0.18\textwidth,frame]{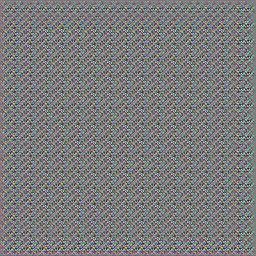}};
    \tikzset{spy on other={
    \node {\pgfimage[width=0.18\textwidth]{figs/imgs/extracted_models/finger_sd14.png}};}}
    \spy on (-1.2,1.2) in node [below] at (1.2,1.8);
    \end{tikzpicture}
    & &

    \begin{tikzpicture}[spy using outlines={rectangle,red,magnification=2.,size=1.5cm, connect spies}]
    \node {\includegraphics[width=0.18\textwidth,frame]{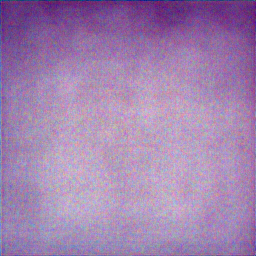}};
    \tikzset{spy on other={
    \node {\pgfimage[width=0.18\textwidth]{figs/imgs/mean_models/finger_glide_fake.png}};}}
    \spy on (-1.2,1.2) in node [below] at (1.2,1.8);
    \end{tikzpicture}
    & 

    \begin{tikzpicture}[spy using outlines={rectangle,red,magnification=2.,size=1.5cm, connect spies}]
    \node {\includegraphics[width=0.18\textwidth,frame]{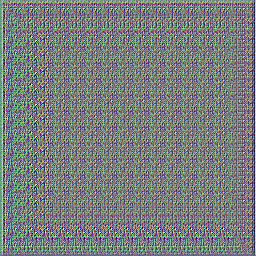}};
    \tikzset{spy on other={
    \node {\pgfimage[width=0.18\textwidth]{figs/imgs/extracted_models/finger_glide.png}};}}
    \spy on (-1.2,1.2) in node [below] at (1.2,1.8);
    \end{tikzpicture}
    \\
    $\mathcal{F}_A$ & $\mathcal{F}_E$ & & $\mathcal{F}_A$ & $\mathcal{F}_E$ \\
    \multicolumn{2}{c}{\textbf{SD 1.4}} & & \multicolumn{2}{c}{\textbf{GLIDE}} 
\end{tabular}
 \caption{
 Fingerprints of SD 1.4 and GLIDE. Per each model two types of fingerprint are shown: fingerprint by residual averaging ($\mathcal{F}_A$) and extracted by DIF ($\mathcal{F}_E$). Observe the $\mathcal{F}_A$ of each model: SD 1.4 demonstrates strong grid-like pattern, whereas GLIDE shows none - GLIDE has a ``weak'' fingerprint. In contrast to $\mathcal{F}_A$, $\mathcal{F}_E$ reveals clear patterns of SD 1.4 and GLIDE.}
 \label{fig:examplesExt}
\end{figure*}%

%% file: figs/tables/comparisonGAN.tex

\begin{table*}[ht]
    \centering
    \footnotesize
    \begin{tabular}{ cclcccccccc }
    \toprule 
    $\boldsymbol{N_P}$ & $\boldsymbol{N_S}$ & \textbf{Method} & \textbf{CycleGAN} & \textbf{StyleGAN} & \textbf{StyleGAN2} &\textbf{StarGAN} & \textbf{BigGAN} & \textbf{GauGAN} & \textbf{ProGAN$_e$} & \textbf{Mean} \\
    \midrule
    \multirow{4}{2.em}{\textbf{0}} & \multirow{4}{2.em}{1024} 
    & Joslin20 & 51.6 & 72.5  & 52.4 & 68.6 & 50.8 & 53.9 & 50.2  & 57.1\\
    & & Marra18 & 58.8 & 82.6 & 50.9  & 92.5 & 54.8 & 51.5 & 48.5 & 62.8\\
    & & Ning18 & 49.5 & 61.7 & 57.2  & 64.1 & 57.7 & 55.5 & 50.2 & 56.6\\
    & & DIF &  \textbf{94.4} & \textbf{96.6} & \textbf{91.5} & \textbf{99.9} & \textbf{96.9} & \textbf{91.8} & \textbf{57.7}  & \textbf{89.8} \\
    \midrule
    
    \multirow{6}{2.em}{\textbf{720k}} & \multirow{3}{2.em}{1024}
    & \RESG  & 91.7 & 94.1 & 94.0  & 95.6 & 85.7 & 95.0 & \textbf{100} & 92.6\\
    & & \RESML  & 93.0 & 94.8 & 92.6  & 98.3 & 93.0 & 95.9 & 95.1 & 94.7\\ 
    & & \RESMG & \textbf{98.0} & \textbf{100} & \textbf{100} & \textbf{100} & \textbf{98.2} & \textbf{98.5} & \textbf{100} & \textbf{98.8}\\
    \cmidrule{2-11}
    
    & \multirow{3}{2.em}{0}
    & \RESG  & 84.6 & 76.5 & 72.2  & 84.7 & 59.4 & 82.9 & \textbf{100} & 91.9\\
    & & \RESML  & 50.6 & 59.8 & 51.2  & 45.7 & 51.9 & 46.3 & 51.2 & 51.0\\ 
    & & \RESMG  &  \textbf{93.5} & \textbf{100} & \textbf{100} & \textbf{99.9} & \textbf{96.5} & \textbf{90.9} & \textbf{99.9} & \textbf{97.2}\\
    \bottomrule
     
    \end{tabular}
    \caption{Classification accuracy (\%). $N_S$ and $N_P$ are the amount of train samples and pre-train dataset size. DIF achieves best result comparing to rule-based methods and is on par with pre-trained models. Low accuracy with DIF for ProGAN$_e$ is expected as it is a mix of images generated from 20 different models.} 
    \label{tab:accuracyGAN}
\end{table*}

%% file: figs/figures/cats_cross_eval.tex
\begin{figure}
    \centering
    \includegraphics[width=1.0\linewidth]{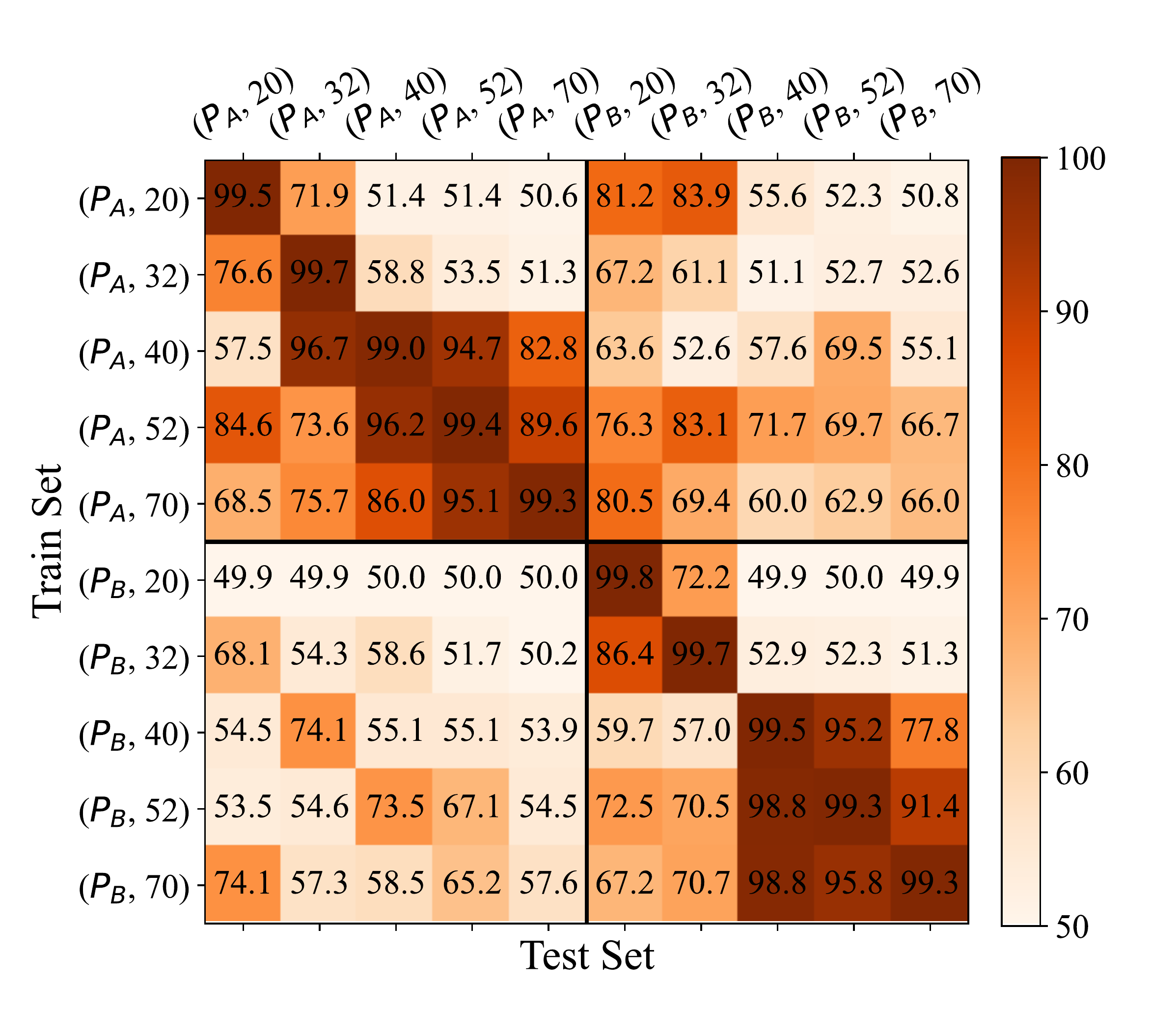}%
    \caption{Cross-detection accuracy in percents. Each grid characterized by model and epoch. Observe clusters for epochs 40,52,70.}
    \label{fig:crossevalCATS}
\end{figure}

%% file: figs/figures/cross_eval_sd.tex
\begin{figure*}
    \captionsetup[subfigure]{labelformat=empty}
    \centering
    \begin{subfigure}{0.29\linewidth}
       \includegraphics[width=1.0\linewidth]{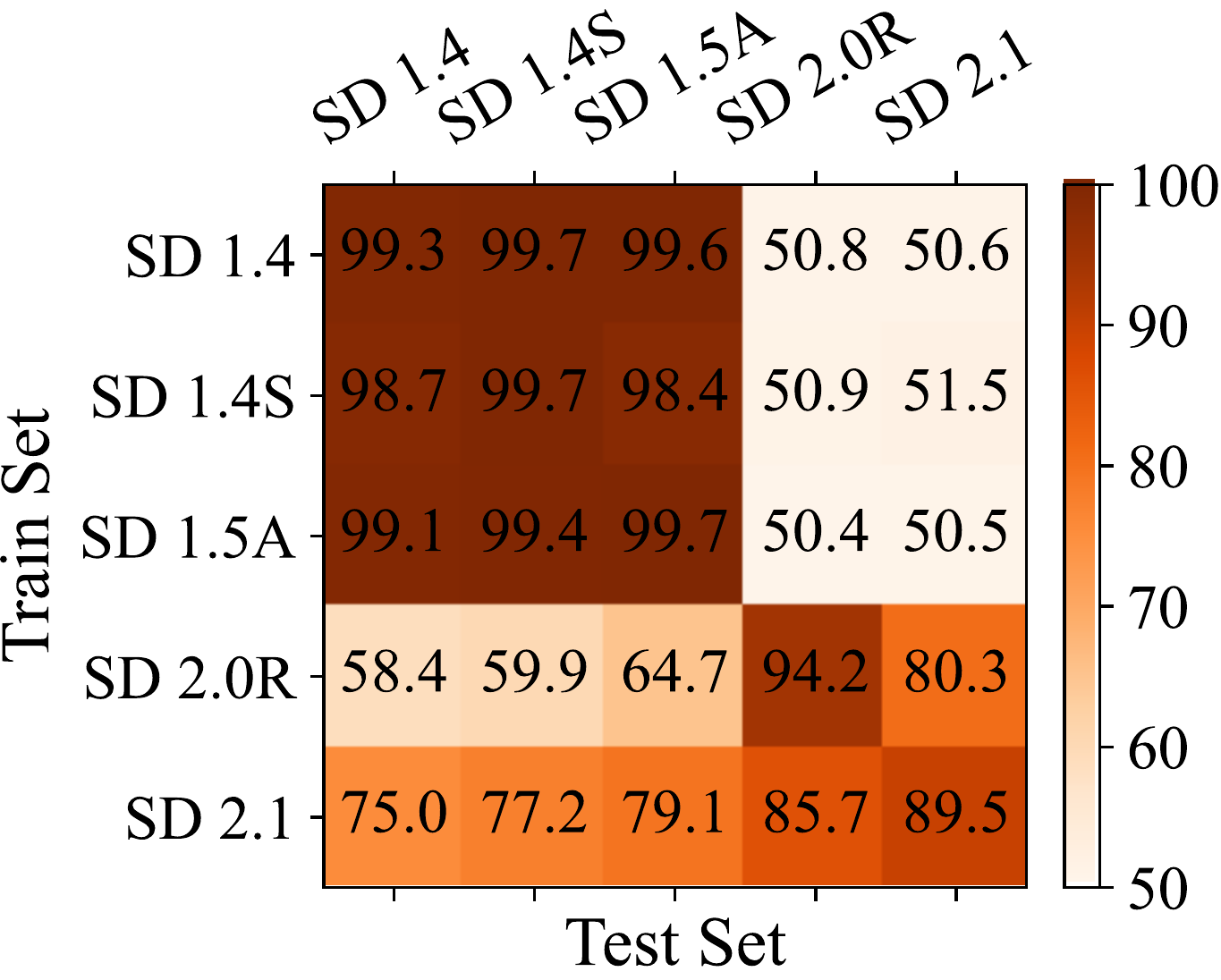}%
      \caption{DIF}
    \end{subfigure}
    \hfill
    \begin{subfigure}{0.29\linewidth}
       \includegraphics[width=1.0\linewidth]{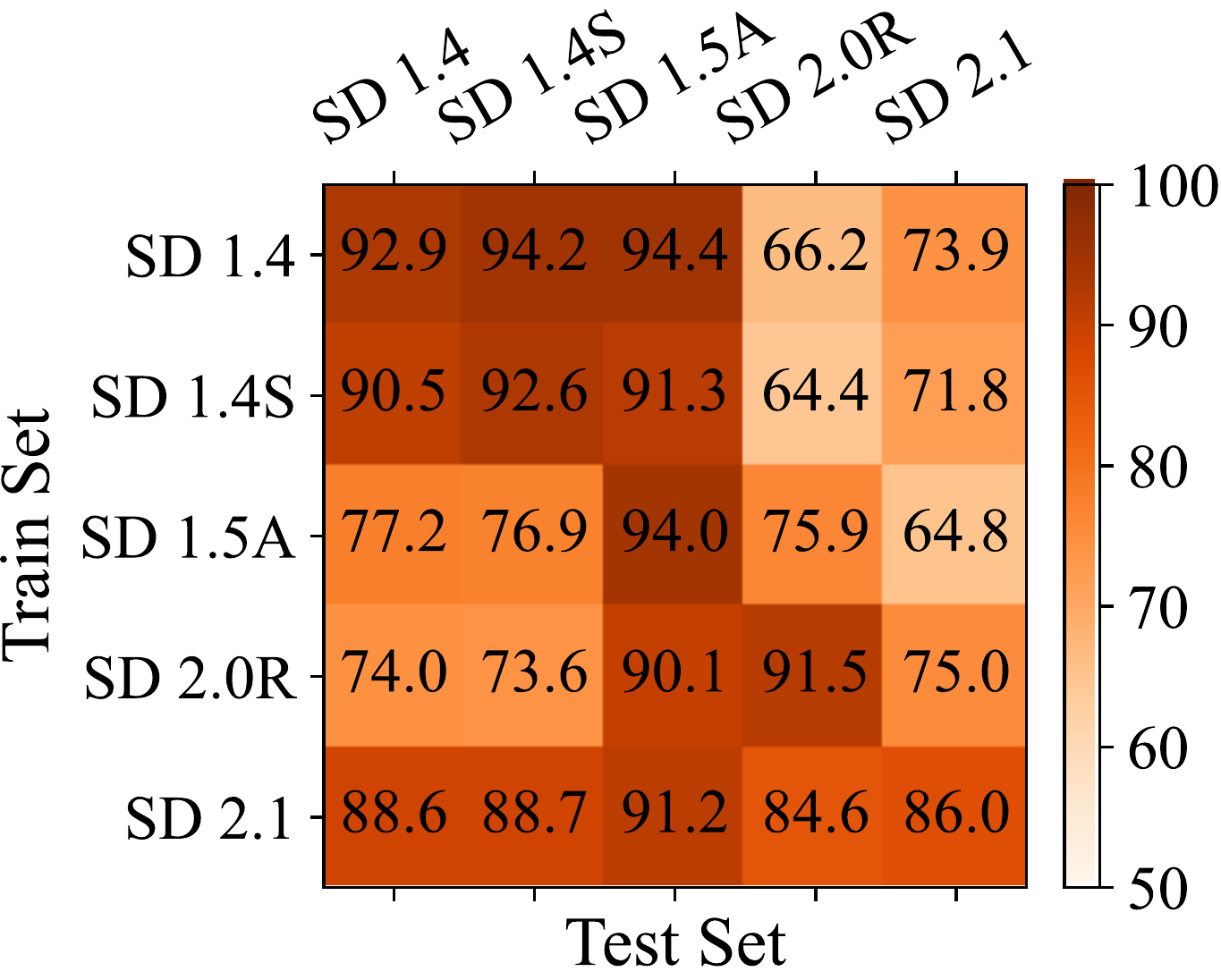}%
      \caption{\RESMG{}}
    \end{subfigure}
    \hfill
    \begin{subfigure}{0.29\linewidth}
		\includegraphics[width=1.0\linewidth]{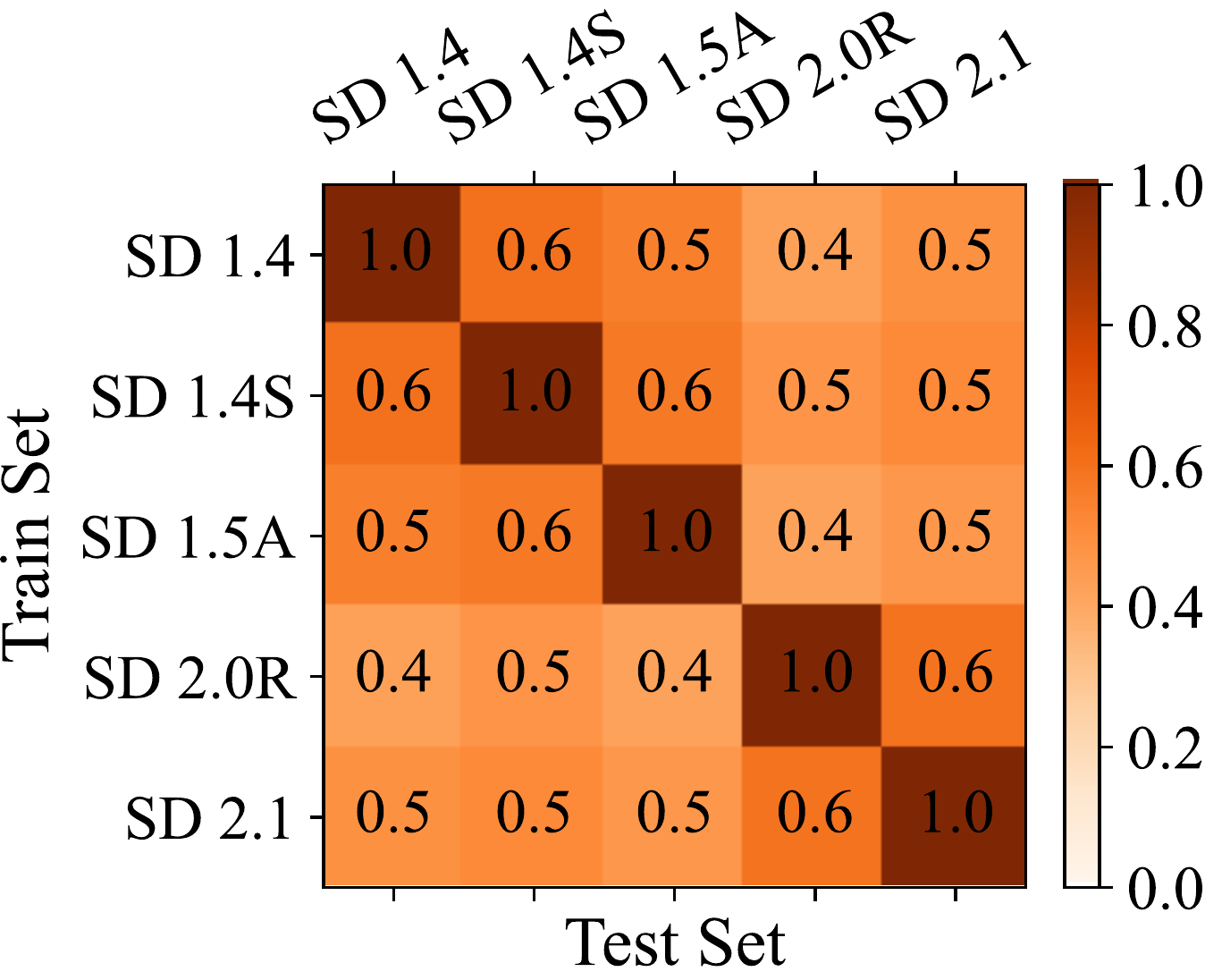}%
      \caption{Marra18}
    \end{subfigure}
    \hfill
    \caption{Model lineage analysis of SD models with different detection methods. For DIF and \RESMG{} we show cross-detection (\%) and for Marra18 cross-correlation of fingerprints. Notably, only DIF exhibits clusters of SD 1.x and SD 2.x with high and symmetric cross-detection.}
    \label{fig:crossevalSD}
\end{figure*}

%% file: figs/figures/cross_eval_t2i.tex
\begin{figure}
    \captionsetup[subfigure]{labelformat=empty}
    \centering
    \begin{subfigure}{0.6\linewidth}
    \includegraphics[width=1.\linewidth]{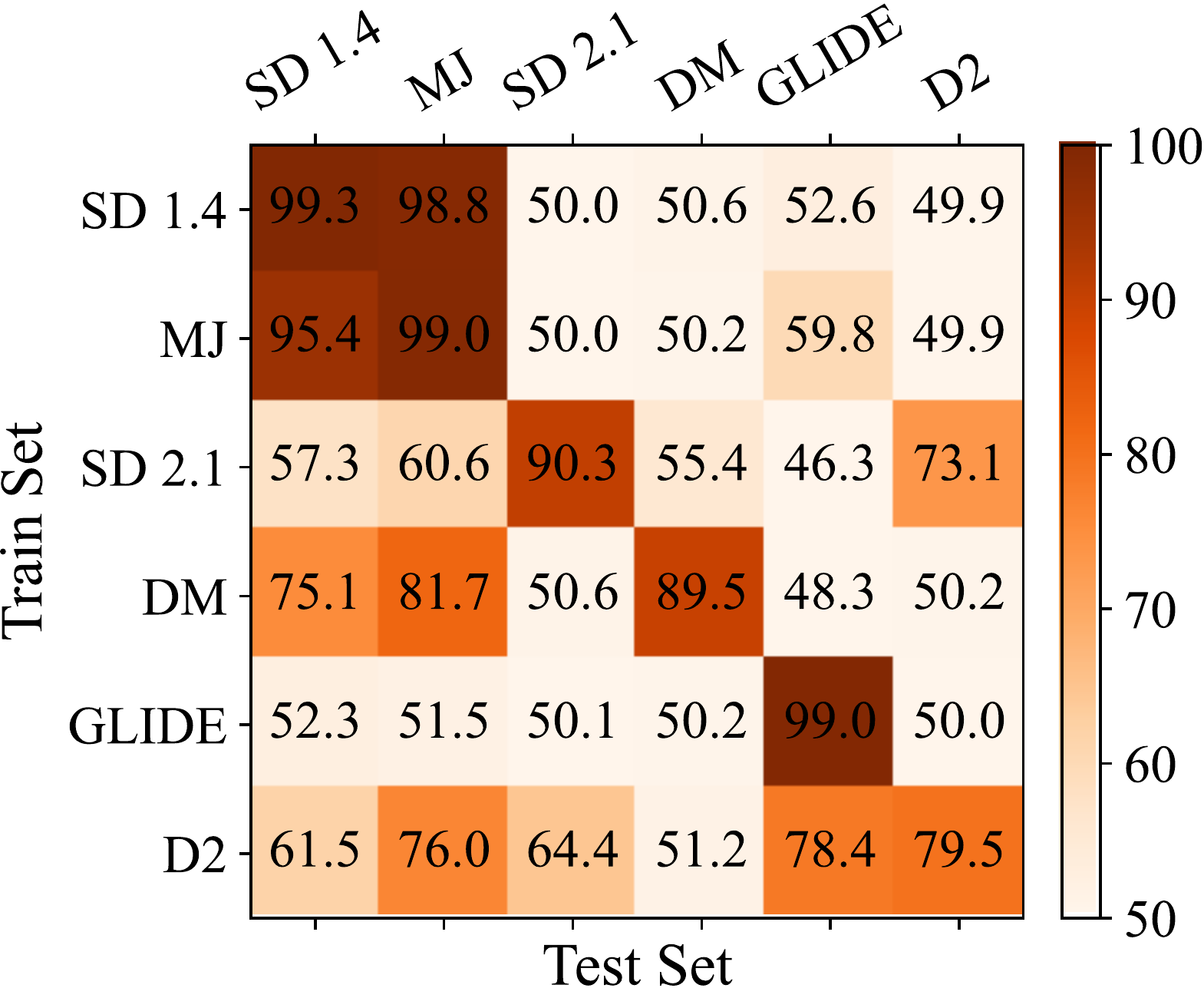}%
    \end{subfigure}
    \caption{Model lineage analysis of \TI{s} with DIF by cross-detection (\%). DM and D2 denote \dalM{} and \dalT{,} respectively. Relation between SD 1.4 and MJ is similar to relation between SD 1.x models.} (\cref{fig:crossevalSD}).
    \label{fig:crossevalTI}
\end{figure}

%% file: figs/figures/compression.tex
\begin{figure}
    \captionsetup[subfigure]{labelformat=empty}
    \centering
    \begin{subfigure}{0.47\linewidth}
      \includegraphics[width=1.0\linewidth]{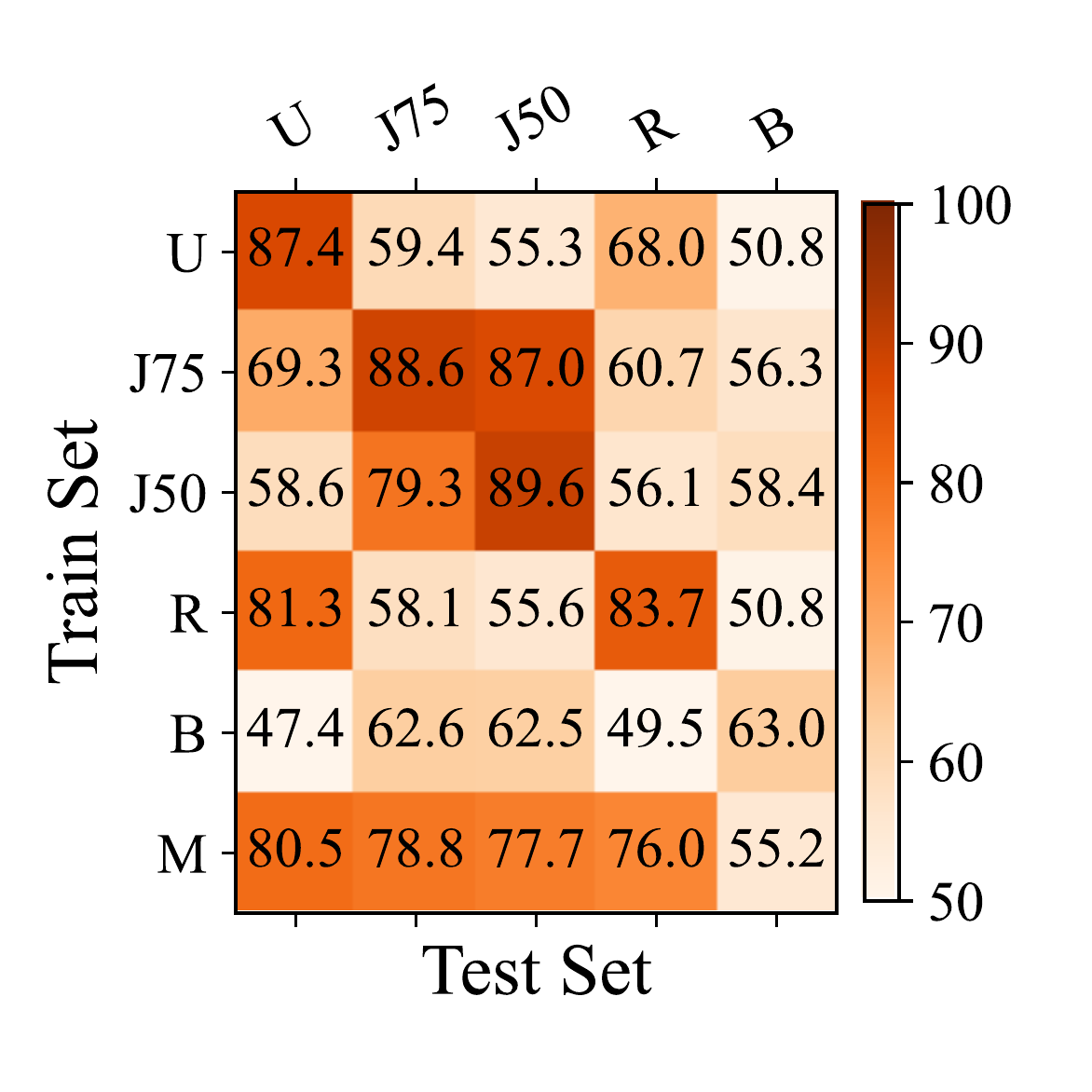}
      \caption{GLIDE}
      \label{fig:compressGLIDE}
    \end{subfigure}
    \hfill
    \begin{subfigure}{0.47\linewidth}
      \includegraphics[width=1.0\linewidth]{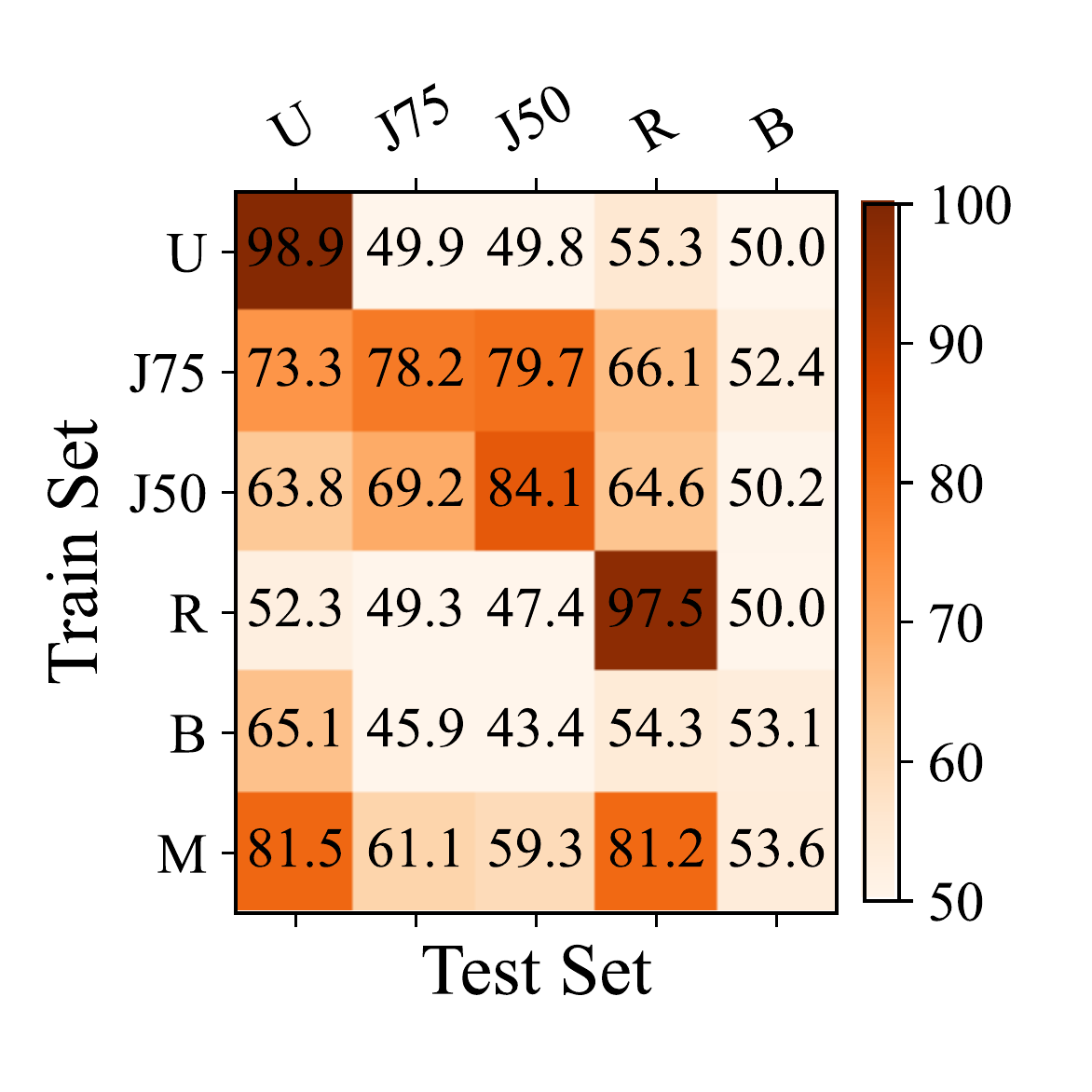}
      \caption{SD 1.4}
      \label{fig:compressSD14}
    \end{subfigure}
      \caption{Detection and cross-detection accuracy per compression for images generated by GLIDE and SD 1.4.}
      \label{fig:compression}
\end{figure}%

%% file: figs/tables/jpeg_ti.tex

\begin{table*}[ht]
    \centering
    \footnotesize
    \begin{tabular}{ cclccccccc }
    \toprule 
    $\boldsymbol{N_P}$ & $\boldsymbol{N_S}$ & \textbf{Method} & \textbf{SD 1.4} & \textbf{SD 2.1} & \textbf{MJ} &\textbf{\dalM}& \textbf{GLIDE} & \textbf{\dalT} & \textbf{Mean}\\
    \midrule
    \multirow{2}{*}{\textbf{0}}  &  \multirow{2}{*}{1024} 
    & \RESM & 72.2 & \textbf{93.9} & 81.1 & 87.9 & \textbf{84.4} & \textbf{73.3} & \textbf{82.1} \\
    & & DIF & \textbf{76.9} & 62.4  & \textbf{83.1} & \textbf{89.8} & 82.7 & 69.7 & 77.4 \\
    
    \midrule

    \multirow{2}{*}{\textbf{720k}} & 1024 
    & \RESML & 99.0 & 98.1 & 98.2 & 93.5 & 94.3 & 79.7 & 93.8 \\
    \cmidrule{2-10}
    & 0 & \RESML & 99.3 & 98.6 & 98.5 & 75.0 & 56.0 & 52.0 & 79.9\\
    \bottomrule
    \end{tabular}
    \caption{Classification accuracy (\%). $N_S$ and $N_P$ are the amount of train samples and pre-train dataset size. Real images and fake images are compressed. DIF achieves slightly worse accuracy relative to \RESM. Pre-trained model \RESML{} shows similar results to uncompressed setting, presumably due to augmentations during training. 
    }
    \label{tab:jpegTI}
\end{table*}

%% file: conclusion.tex
\section{Conclusions and Future Work}\label{sec:conclusion}

This study provides a twofold contribution: the development of new methods for synthetic image detection and the establishment of a methodology for model lineage analysis. We have shown that CNNs naturally exhibit image artifacts, which we leverage in our method called DIF. DIF extracts fingerprints from generated images, allowing us to detect images that come from the same model or its fine-tuned versions. Our method achieves high detection accuracy, surpassing methods trained under the same conditions and performing similarly to pre-trained state-of-the-art detectors for generated images from popular models. Remarkably, we achieve these results using a small number of generated images (up to 512), while other detectors require significantly more samples for training.

In terms of model lineage analysis, we employ cross-detection as a means to trace fine-tuned generative models. Notably, our analysis reveals that MidJourney is indeed a fine-tuned variant of the Stable Diffusion 1.x model.

However, we identified several drawbacks that require further investigation. Some image generators produce weak fingerprints, which are challenging for DIF. Furthermore, the method performs poorly on certain compression methods and blurred images. We believe that the effect of compression on fingerprint extraction and detection requires additional attention and is a topic for future research.

%% file: supp_arxiv.tex
\twocolumn[ \centering
\section*{Supplemental~Material}
\vspace{1cm}
]

\renewcommand{\thetable}{\Alph{table}}
\renewcommand{\thefigure}{\Alph{figure}}

\input{supp_sections/supp_method}

\input{supp_sections/supp_detect}

%% file: supp_sections/supp_method.tex
\section{Implementation Details}\label{apx:method}

This section includes the implementation details of our work. The DIF models were trained and tested on an RTX 3060 GPU with 12 GB of VRAM. We used PyTorch~\cite{art:pytorch} as our deep learning framework.

\subsection{Selection of Denoising Filter}

Here we explain why DnCNN~\cite{art:dncnn} was chosen as the denoising filter (\HP{}) for all the fingerprint methods. The objective of the \HP{} is to extract only noise and artifacts while filtering out the semantic information of the image. The pattern of $\mathcal{F}$ is primarily present in the high-mid frequencies~\cite{art:ganfingerprint}, making a high-pass filter a straightforward choice as a reference \HP{}. However, some semantic content, such as edges, also exists in these frequencies. Therefore, a more sophisticated denoising filter is required. DnCNN is such a filter. It is trained to extract Gaussian noise while preserving edges, and it has demonstrated good performance with other types of noise as well~\cite{art:dncnn}. To validate our assumption, we compare the performance of Marra18, Joslin20, and DIF using both a Gaussian high-pass filter and DnCNN (\cref{tab:filters}).  
\input{figs/tables/filter_comp}

As expected, all of the methods perform worse with the Gaussian high-pass filter. The high-frequency content of the image influences the averaged and extracted fingerprint pattern, leading to low correlation values between unseen residuals (test set) and fingerprints.

\subsection{Usage of DnCNN}

The DnCNN-S~\cite{art:dncnn} model was trained separately from the residual extraction procedure. The training was performed according to the original work with minor changes. We trained the DnCNN-S model for 2,000 epochs with a learning rate of $10^{-4}$ and the Adam optimizer~\cite{art:adam}. Only real images were used during training. Random crop is set to size $(48 \times 48)$ pixels and a sigma range is set to [5, 15]. The number of training images is 1024.

During the inference of the DnCNN, we applied post-processing. First, the input to the DnCNN was padded with 10 pixels on each dimension and then reduced as post-processing according to the recommendations of the authors~\cite{art:dncnn}. Additionally, we observed a bias within the residuals, so we performed an additional post-processing step. We took the training set of the DnCNN and calculated the average of its residuals, thus estimating the fingerprint of the DnCNN ($\mathcal{F}_{DnCNN}$). During the inference of the model, this fingerprint was subtracted from the output according to:
\begin{equation}
    R_i = f_{D}(X_i) - \mathcal{F}_{DnCNN}
\end{equation}

\subsection{U-Net Architecture}\label{sec:}

In \cref{tab:unet}, we summarize the architecture of the U-Net model, denoted in the paper as $g_{\theta}$. Each row represents a convolution block, comprised of two convolutional layers and either an up-sampling or a down-sampling layer. The convolutional layers have a kernel size of 3, stride of 1, and padding of 1 pixels for each spatial dimension to achieve boundary artifacts. Each convolutional layer is accompanied by Batch-Normalization~\cite{art:batchnorm} and an activation function, which are specified in \cref{tab:unet}. We use max-pooling with a kernel size of $2 \times 2$ pixels for down-sampling and a deconvolution layer with a kernel size of $2 \times 2$ pixels and stride of 2 pixels for up-sampling. The latter is suspected to be the main causes of grid-like artifacts~\cite{art:checkerboard}. The last block consists only of a single convolution layer where we do not use Batch-Normalization and rely on hyperbolic tangent (TanH) as the activation function.
\input{figs/tables/unet}

\subsection{Selection of Model Architecture}

We selected the U-Net architecture through hyperparameter tuning and by incorporating concepts mentioned in Section 3.1. A Convolutional Network (C-Net) consists of convolutional blocks without down-sampling and up-sampling operations. An Up-Sampling Network (D-Net) serves as the decoder in the U-Net model. U-Net with 1x1 kernels within convolutional layers denoted as U1-Net, where we test the importance of only up-sampling artifacts. U-Net refers to the aforementioned architecture.

In \cref{tab:accuracyArch}, U-Net outperforms other architectures, while C-Net shows the worst performance due to the lack of up-sampling, which is crucial (Section 3). Interestingly, C-Net shows minimal decline when used with the GLIDE model, indicating a dominant presence of boundary artifacts (Figure 2). Additionally, U1-Net, designed to reduce boundary artifacts, experiences approximately a 5\% decline compared to other architectures when used with GLIDE.

D-Net performs similarly to U-Net as both models generate both up-sampling and boundary artifacts. However, U-Net was chosen due to its superior performance.

\input{figs/tables/arch_compare}

%% file: figs/tables/filter_comp.tex
\begin{table}[ht]
    \centering
    \small
    \begin{tabular}{ lcc }
    \toprule 
    \textbf{Method} & \textbf{Gaussin High-Pass} & \textbf{DnCNN}\\
    \midrule
    Marra18  & 52.4 & \textbf{62.6} \\
    Joslin20  & 51.0 & \textbf{51.7} \\
    DIF  & 73.3 & \textbf{92.3}  \\
    \bottomrule
    \end{tabular}
    \caption{Mean accuracy (\%) of detection for \TI{s}. 1024 images were used in train set for each. 
    \label{tab:filters}}
\end{table}

%% file: figs/tables/unet.tex
\begin{table}
    \centering
    \small
    \begin{tabular}{ p{4em}ccc }
    \toprule 
    \textbf{Model}&  $\boldsymbol{C_{in}}$  & $\boldsymbol{C_{out}}$ & $\boldsymbol{f_{act}}$ \\
    
    \midrule
    
    \multirow{4}{2.em}{\textbf{Encoder}} &  16  & 32 & Leaky-ReLU \\
      & 32  & 64 & Leaky-ReLU  \\
      & 64  & 128 & Leaky-ReLU  \\
      & 128  & 256 & Leaky-ReLU  \\
    
    \midrule
    
    \multirow{5}{2.em}{\textbf{Decoder}} &  256+256  & 128 & Leaky-ReLU  \\
      & 128+128  & 64 & Leaky-ReLU  \\
      & 64+64  & 32 & Leaky-ReLU  \\
      & 32+32  & 32 & Leaky-ReLU  \\
      & 32  & 3 & TanH  \\
    
     \bottomrule 
     
    \end{tabular}
    \caption{U-Net model architecture. Each row represents a convolutional block, with $C_{in}$ and $C_{out}$ indicating the input and output channels, respectively. $f_{act}$ is the activation function used in the block. The symbol ``+'' in $C_{in}$ indicates that the layer input includes skip connections. 
    \label{tab:unet}}
\end{table}

%% file: figs/tables/arch_compare.tex
\begin{table*}[ht]
    \centering
    \small
    \begin{tabular}{ lcccccc|c }
    \toprule 
    \textbf{Architecture} & \textbf{SD 1.4} & \textbf{SD 2.1} & \textbf{MJ} & \textbf{\dalM} & \textbf{GLIDE} & \textbf{\dalT}  & \textbf{Mean}\\
    \midrule
    U-Net & 99.3 & \textbf{89.5}  & \textbf{99.0} & \textbf{99.0} & \textbf{90.3} & \textbf{79.5}  & \textbf{92.8}\\
    U1-Net & 99.4 & 88.8  & 98.9 & 98.3 & 85.2 & 80.2 & 91.8\\
    D-Net  & \textbf{99.4} & 89.3  & 98.9 & 99.0 & 89.0 & 78.0 & 91.9\\
    C-Net  & 60.7 & 53.3  & 89.9 & 98.0 & 88.2 & 69.6 & 76.6\\
     \bottomrule 
    \end{tabular}
    \caption{Detection accuracy (\%) of CNN architectures on images from \TI{s}. D-Net is a up-sampling network and C-Net convolutional network. U-Net demonstrates best results.
    \label{tab:accuracyArch}}
\end{table*}

%% file: supp_sections/supp_detect.tex
\section{Datasets}\label{apx:data}

\subsection{GAN Datasets}

The GAN datasets that we use in this work are from the supplementary materials of Wang~\etal~\cite{art:easycnn}. Train set consists of 360k real images corresponding to 20 LSUN classes~\cite{art:lsun} and 360k images generated by 20 ProGAN~\cite{art:progan} models. Models are trained per LSUN class. For the ProGAN dataset we randomly select 2,000 images per real and fake class, for each LSUN image class, resulting in 4,000 images per LSUN class overall. Test set of~\cite{art:easycnn} contains images generated by a number of GANs, specifically by StyleGAN~\cite{art:stylegan}, StyleGAN2~\cite{art:stylegan2}, BigGAN~\cite{art:biggan}, StarGAN~\cite{art:stargan}, GauGAN~\cite{art:gaugan} and ProGAN~\cite{art:progan}.

\subsection{Fined-Tuned Stable Diffusion Models}

We will outline the process of constructing datasets for the fine-tuned Stable Diffusion models discussed in \cref{sec:source}.
In all cases, fine-tuning involves the DreamBooth method~\cite{art:dreambooth}, specifically \TI{} is trained to reproduce a target object or style from a pre-defined keyword. There are three models: SD 1.4S, SD 1.5A, and SD 2.0R. The authors customly fine-tuned SD 1.4S with 10 photos. SD 1.5A\footnote{\niceurl{https://huggingface.co/DGSpitzer/Cyberpunk-Anime-Diffusion}}
is a publicly available model that was fine-tuned in two stages: first, SD 1.4 was fine-tuned with 680k anime images, and then, after replacing the image decoder with one from SD 1.5, it was additionally fine-tuned with~\cite{art:dreambooth} on a different set of anime images. The last model, SD 2.0R\footnote{\niceurl{https://huggingface.co/nousr/robo-diffusion-2-base}}%
, is fine-tuned with an unspecified amount of robot images. Following above, keywords are known for each case, as SD 1.4S is trained by us, and for others, keyword is specified in the instructions for each model. To produce images with a new models a keyword is added to captions from the \laion{} dataset~\cite{misc:laion5b} in the following format: ``keyword caption''. This forces fine-tuned model to generate images with new information previously unknown to the source models. \cref{fig:sdExamples} shows an example of outputs resulting from the same caption.

\section{Additional Results}\label{apx:adresults}

\input{figs/figures/cross_eval_gans}
\subsection{The Effect of Train Set Size}

In Section 4 
we present an evaluation of the proposed method as a detector of generated images. To evaluate the method under various amounts of training samples, we measure performance on images produced by both GAN and \TI{} models, as shown in \cref{fig:trainGAN,fig:trainT2I}. We observe that the accuracy remains stable across most of the models and starts to degrade drastically at $N_S = 128$, where we observe a 5\% drop for the majority of the models.

\input{figs/figures/gan_train_plot}

\subsection{Consistency for a Low Sample Amount}

We report the consistency of accuracy with a low number of samples. In this setting, each training sample has a greater impact on the results. To test this effect, we trained four models (\RES, \RESMG, \RESM, and DIF) ten times with 128 samples according to Section 4.2. Each time we randomly sampled the training set. As shown in \cref{fig:boxes}, despite the weaker consistency of the results for DIF, statistics support the relation observed in Table 2.

\input{figs/figures/box_plots}
\input{figs/figures/sd_examples}
\subsection{Cross-Detection for GAN Models}

We report the complete map of cross-detection for datasets corresponding to GAN and ProGAN$_t$ models in \cref{fig:crossevalGAN,fig:crossevalPROGANA} respectively. As shown in the figures, the cross-correlation is low for all the models, thus models were trained separately.
\input{figs/figures/cross_eval_progans}
\subsection{Cross-Detection for Custom Trained ProGANs}

Models $P_A$, $P_B$, $\hat{P}_A$ and $\hat{P}_B$ were trained on centrally cropped images ($128 \times 128$ px) from AFHQ~\cite{art:stargan2}. We report the full cross-detection matrix for the four models in \cref{fig:crossevalCUSTOMF}. The cross-detection across all the models is similar, namely no symmetry between different models and high-symmetry within converged models. In addition we present the examples of images generated by $P_A$ ordered according to check point epochs at \cref{fig:catExamples}. Image quality corresponds to the convergence state of the model. Starting from epoch 40 the model produces visually appealing results, that correspond to symmetric and high cross-detection accuracy observed in \cref{fig:crossevalCUSTOMF}. Cross-detection for $\hat{P}_A$ is symmetric, but with lower cross-detection accuracy values. We suspect that in this specific case, the model did not converge properly.

\subsection{Cross-Detection vs. Cross-Correlation}

In this section, we clarify why model lineage estimation is performed using image cross-detection rather than fingerprint cross-correlation. Cross-correlation (Section 3) is an intuitive process where we expect a correlation value of 1 for a fingerprint matched with itself, high absolute values for similar fingerprints, and low values for unrelated fingerprints. However, in our experiments (Section 4.3), we found that these values are not informative. The absolute cross-correlation values of the extracted fingerprints ($\mathcal{F}_E$) exhibit a random pattern (\cref{fig:crossevalCUSTOMFE}), while the fingerprints obtained through residual averaging ($\mathcal{F}_A$) display a trend similar to cross-detection (\cref{fig:crossevalCUSTOMFA}). However, the cross-detection of $\mathcal{F}_A$ decreases more rapidly with epochs, lacks normalization, and is always symmetric. As a result, the certainty of model relationships is diminished due to: a) increased sensitivity to changes during training/fine-tuning, b) absence of a symmetry parameter, and c) insufficient information from the correlation value alone.

\input{figs/figures/custom_gans_full}

\input{figs/figures/custom_gan_fingers}

%% file: figs/figures/cross_eval_gans.tex
\begin{figure}
    \centering
    \includegraphics[width=1.\linewidth]{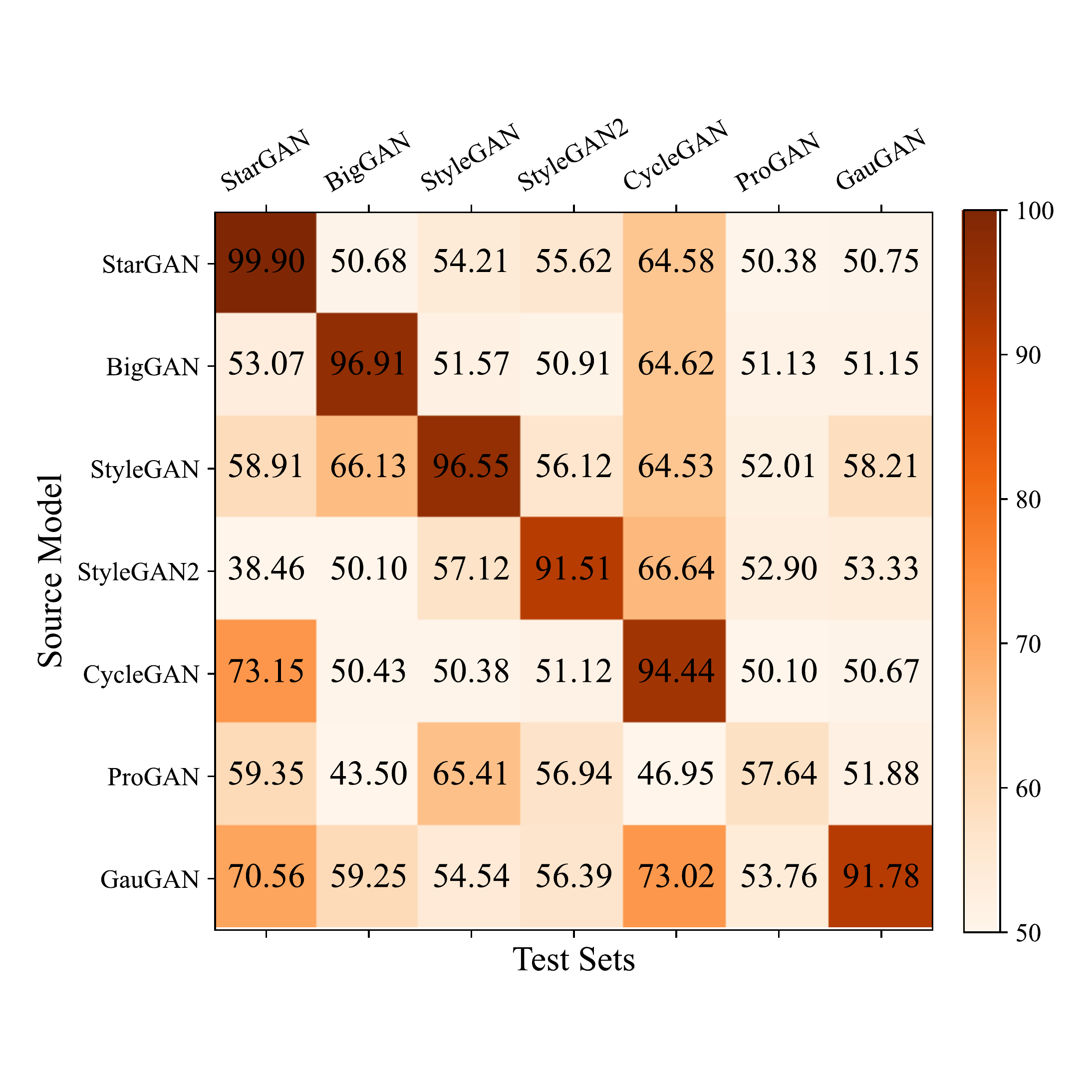}%
    \vspace*{0em}
    \caption{Cross-detection accuracy (\%) for GAN models.}
    \label{fig:crossevalGAN}
\end{figure}

%% file: figs/figures/gan_train_plot.tex
\begin{figure}[h]
    \centering
      \includegraphics[width=\linewidth]{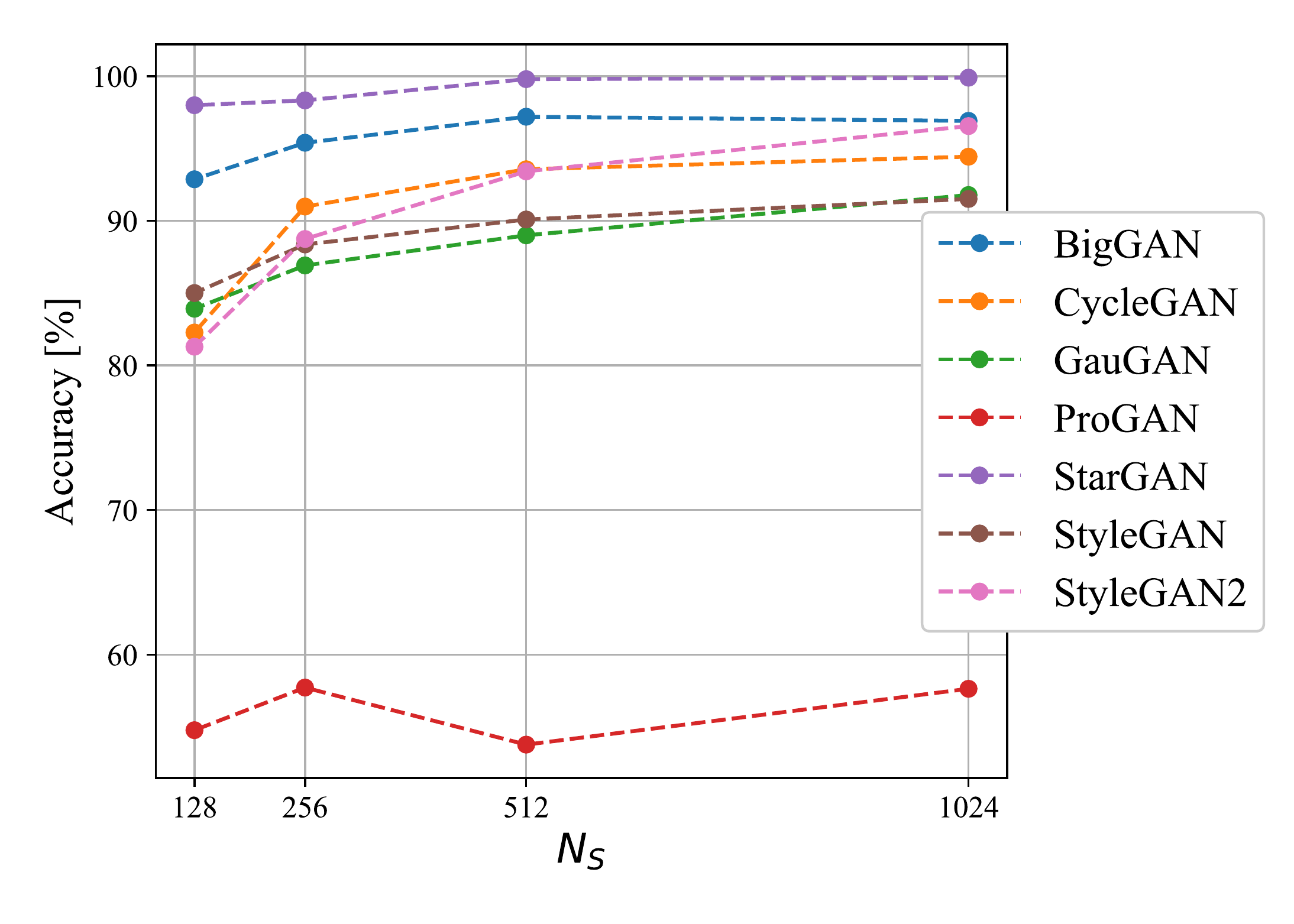}
      \caption{Accuracy (\%) as a function of train samples for GAN models.}
      \label{fig:trainGAN}
\end{figure}%
\begin{figure}[h]
    \centering
      \includegraphics[width=\linewidth]{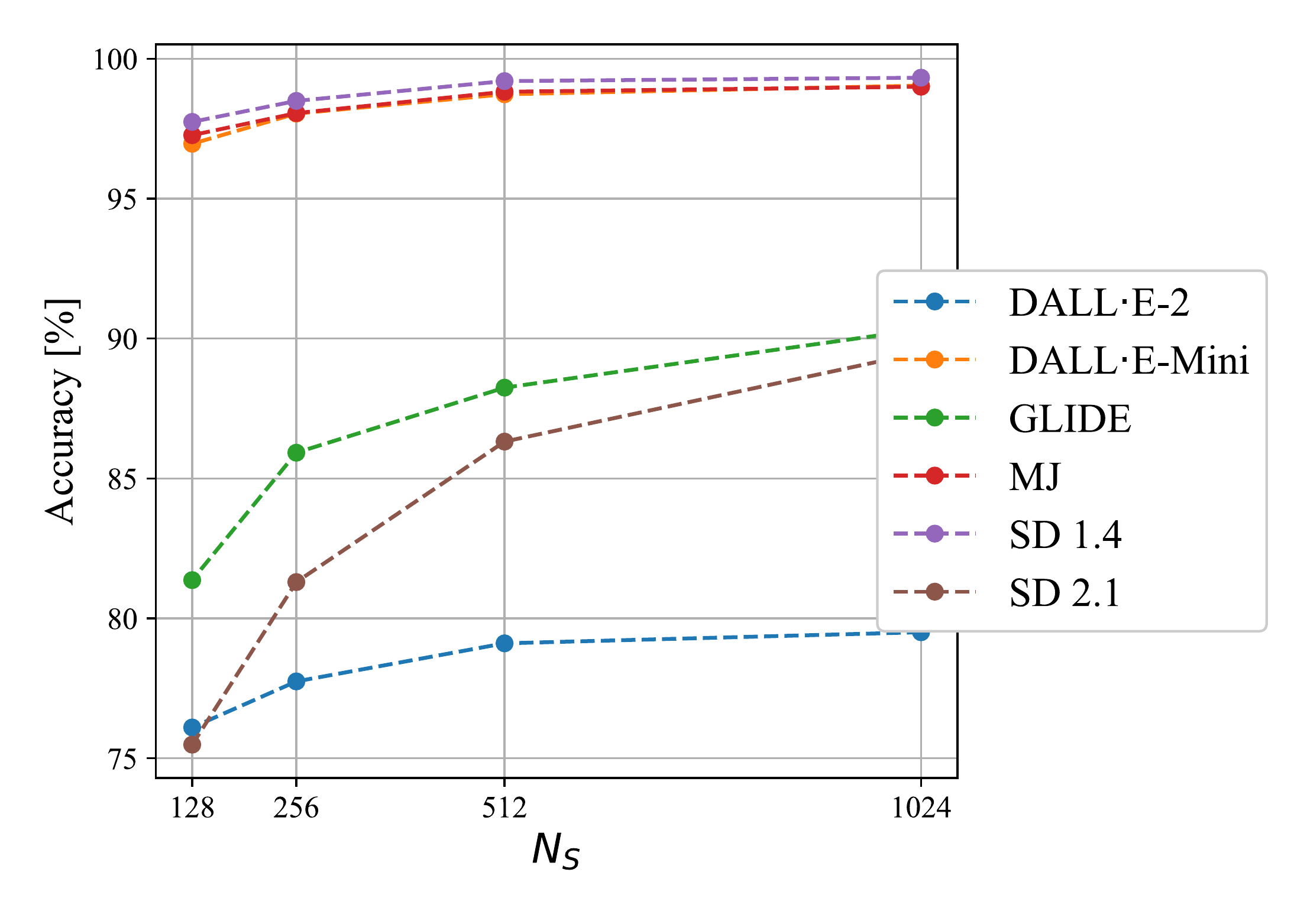}
      \caption{Accuracy (\%) as a function of train samples for \TI{} models.}
      \label{fig:trainT2I}
\end{figure}%

%% file: figs/figures/box_plots.tex
\begin{figure*}[t]
\centering
\setlength{\tabcolsep}{2pt}
\begin{tabular}{cc}
    \includegraphics[width=0.4\textwidth]{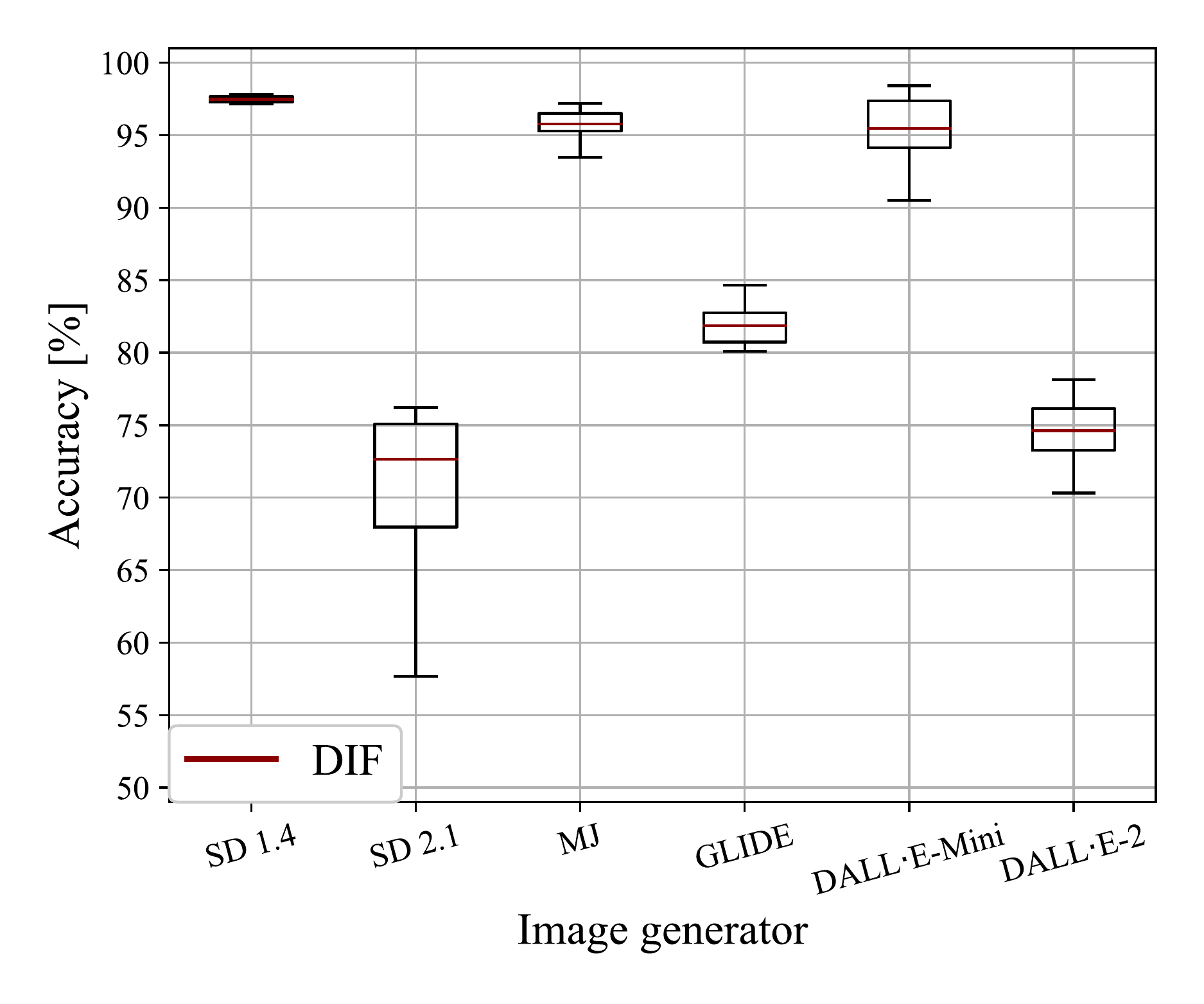}%
    & 
    \includegraphics[width=0.4\textwidth]{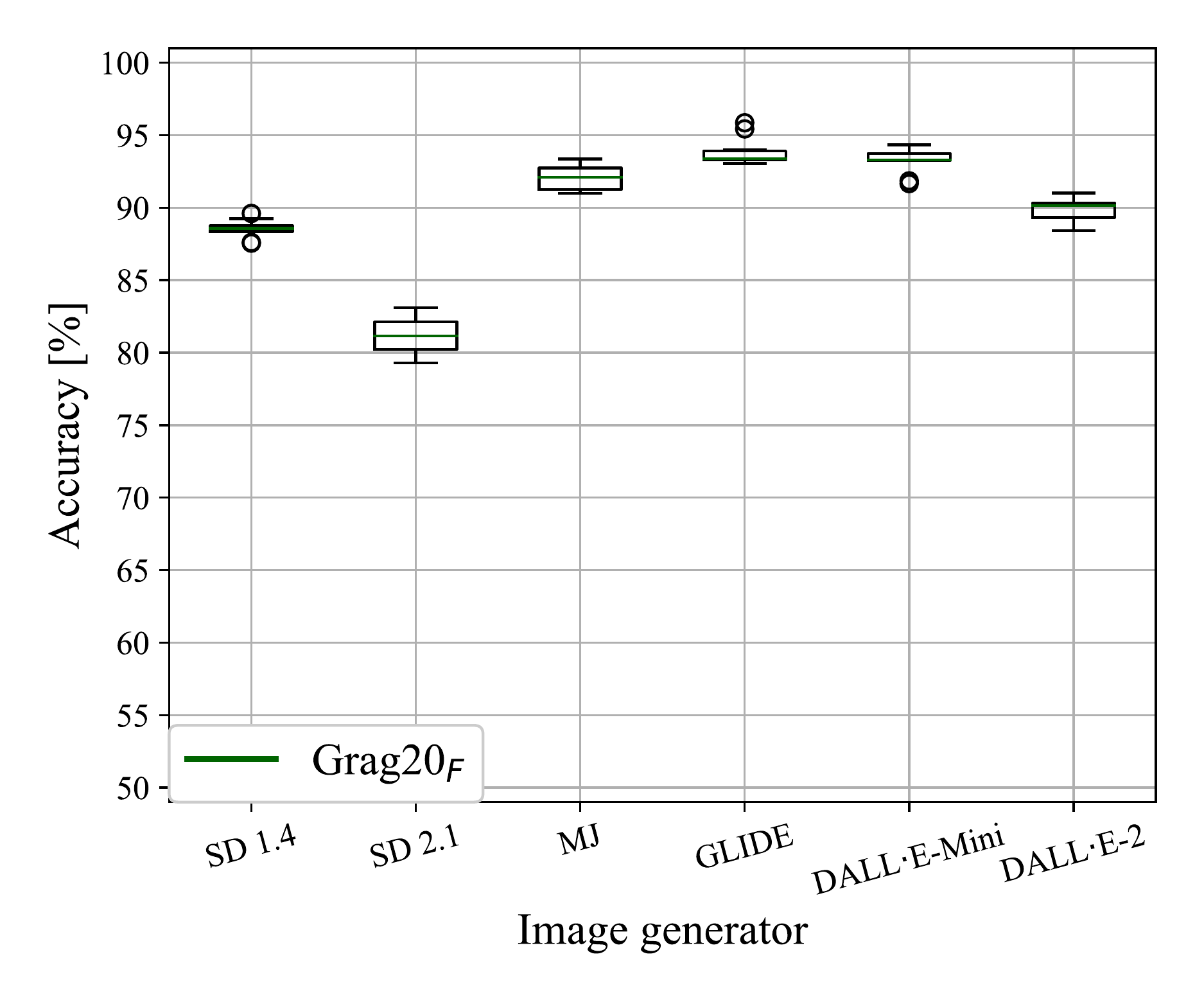}%
    \\%
    \includegraphics[width=0.4\textwidth]{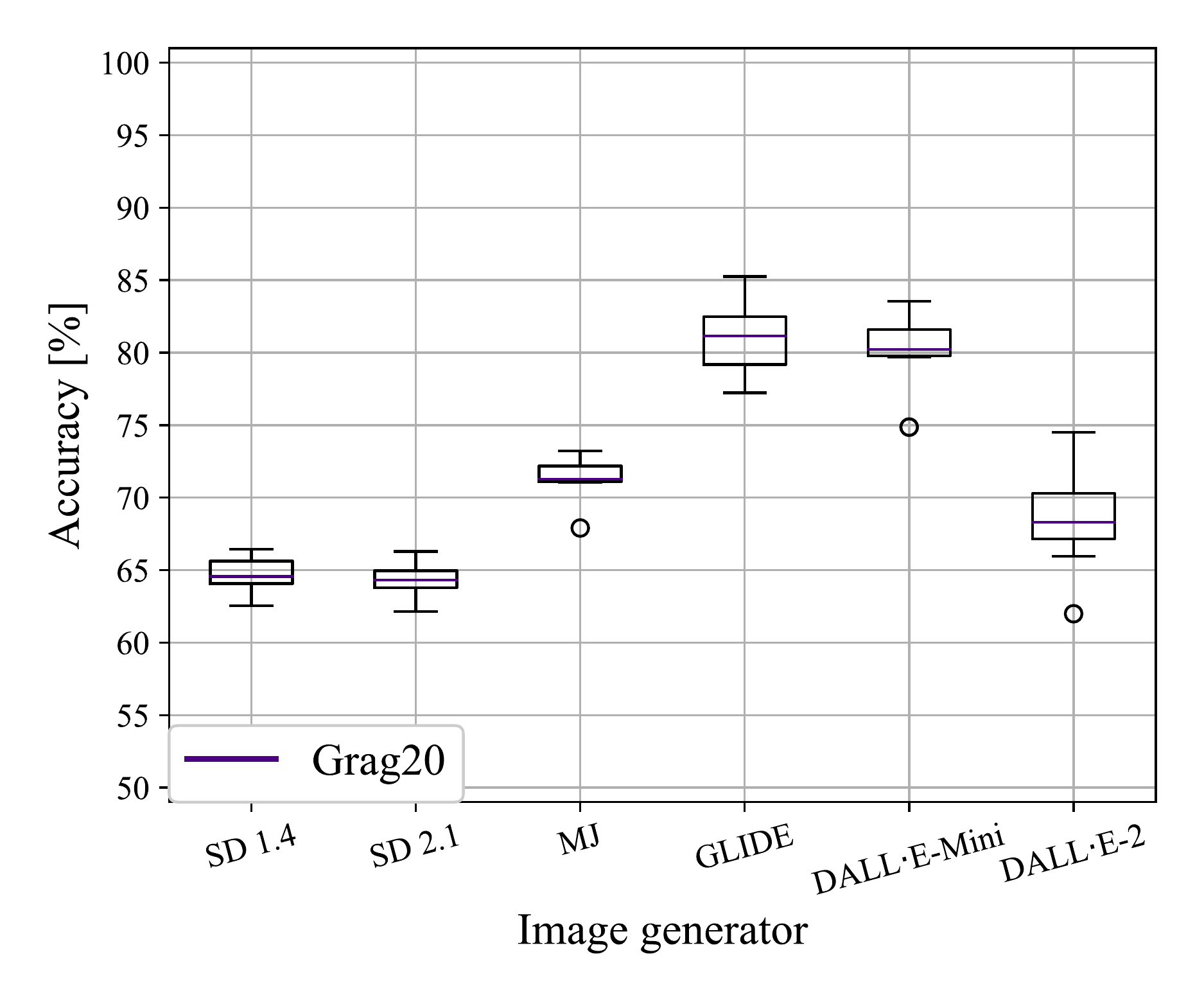}%
    & 
    \includegraphics[width=0.4\textwidth]{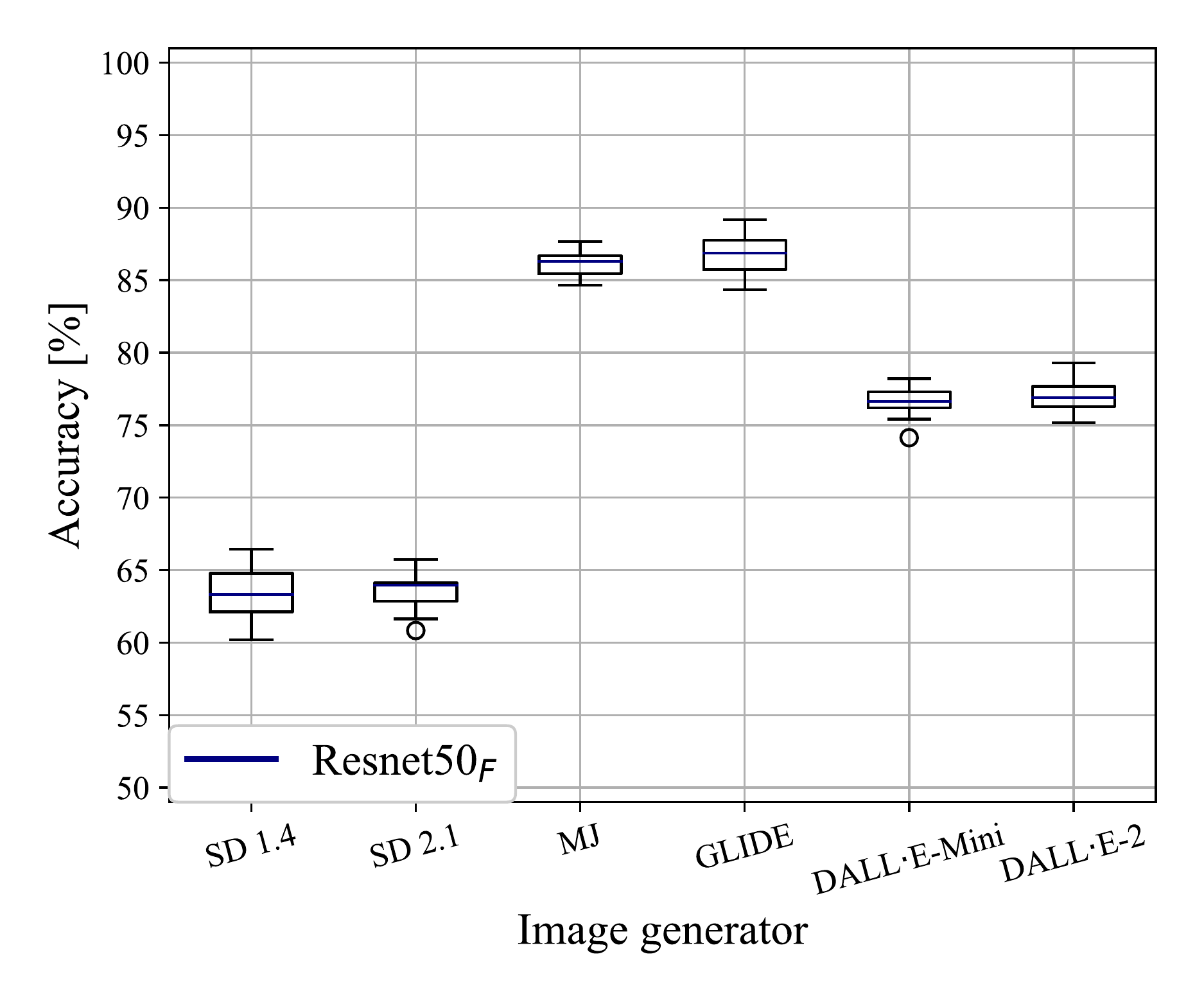}%
    \end{tabular}
    \caption{Box plots of accuracy per image generator dataset. DIF demonstrates less consistent results, especially with SD 2.1., but preserves relation from Table 2.}
    \label{fig:boxes}
\end{figure*}

%% file: figs/figures/sd_examples.tex
\begin{figure*}[t]
\setlength{\tabcolsep}{2pt}
\begin{tabular}{ccccc}
    \centering
    \includegraphics[width=0.19\textwidth,frame]{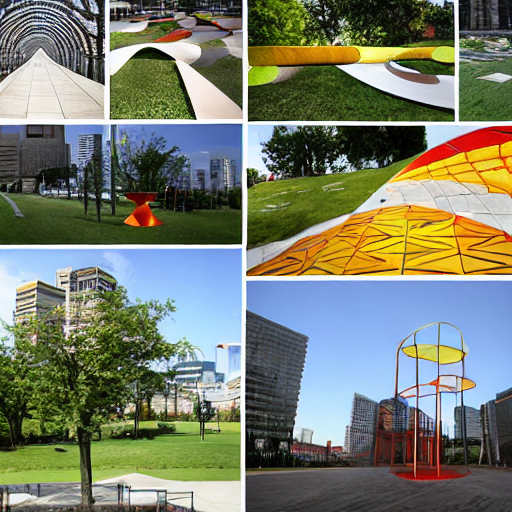}%
    & 
    \includegraphics[width=0.19\textwidth,frame]{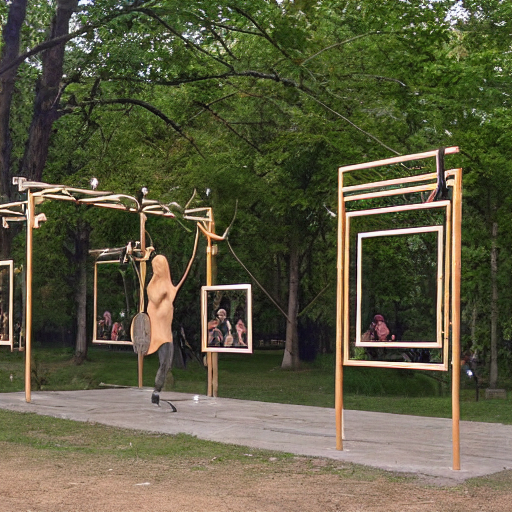}%
    & 
    \includegraphics[width=0.19\textwidth,frame]{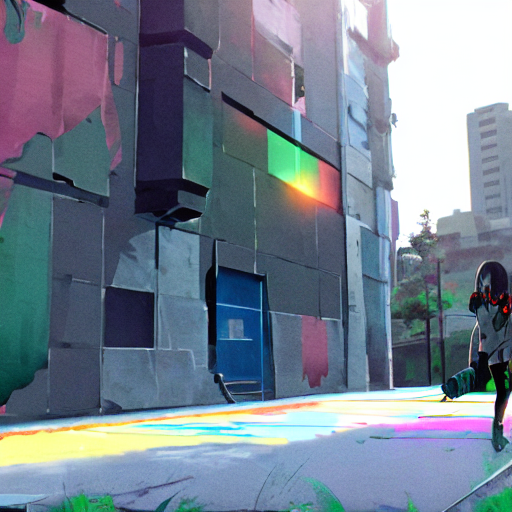}%
    & 
    \includegraphics[width=0.19\textwidth,frame]{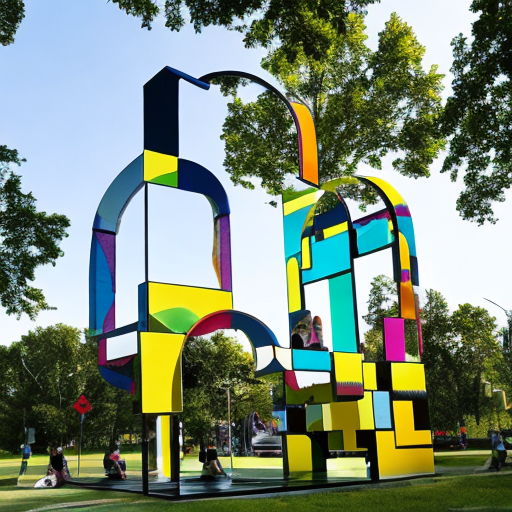}%
    & 
    \includegraphics[width=0.19\textwidth,frame]{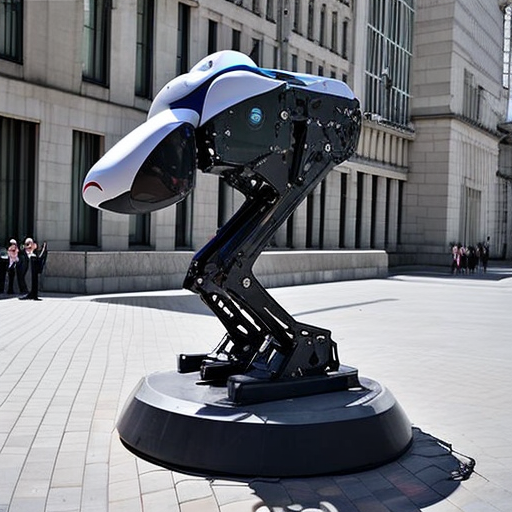} 
    \\%
    SD 1.4 & SD 1.4S & SD 1.5A & SD 2.1 & SD 2.0R
    \end{tabular}
    \caption{Examples of 
    images produced by Stable Diffusion models and their variants. Images correspond to the caption ``Best Public Arts Installations park 2015'' and addition of model specific keywords.}
    \label{fig:sdExamples}
\end{figure*}

%% file: figs/figures/cross_eval_progans.tex
\begin{figure*}[t]
    \centering
    \includegraphics[width=1.\textwidth]{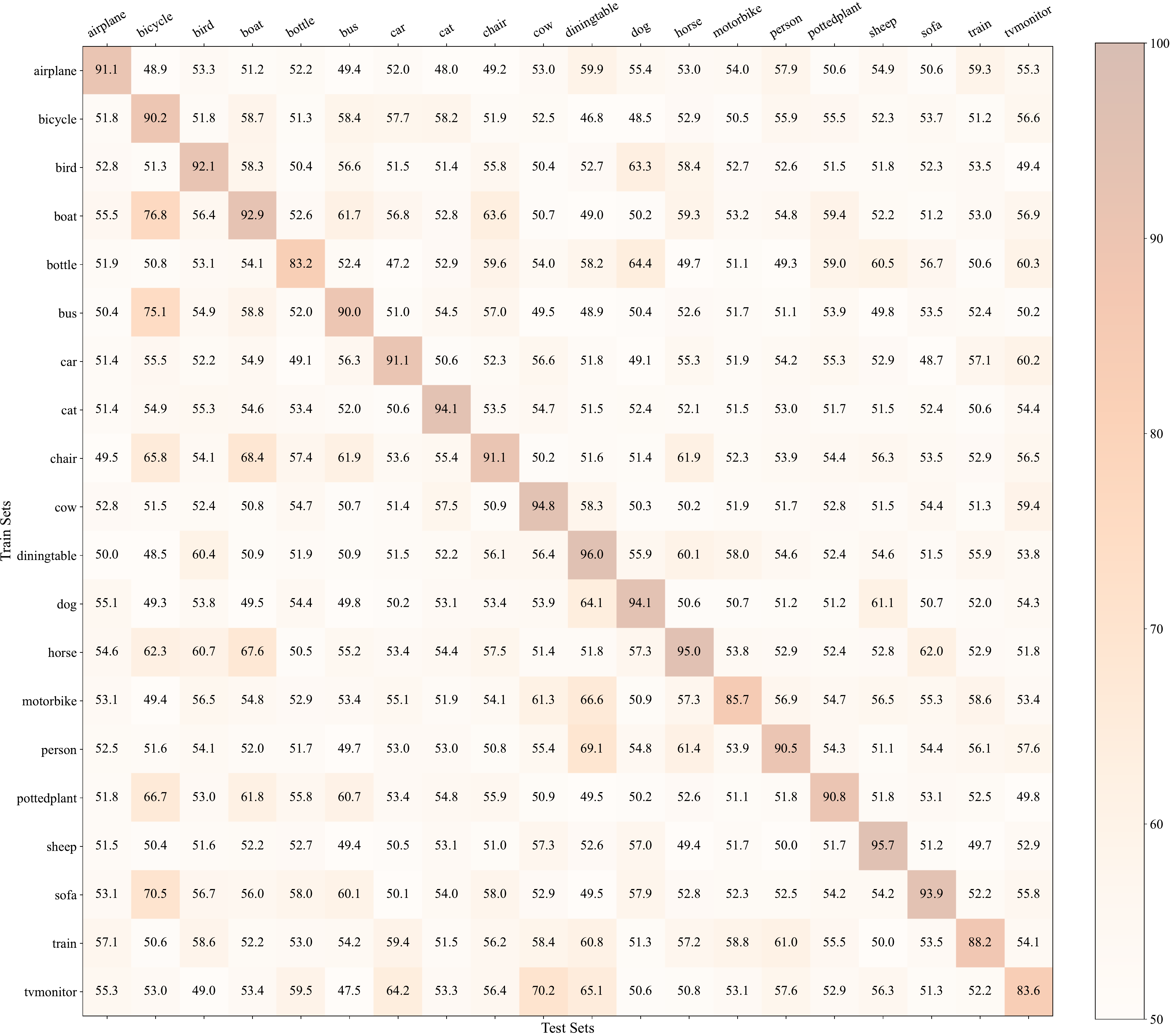}%
    \caption{Accuracy (\%) cross-detection for ProGAN models.}
    \label{fig:crossevalPROGANA}
\end{figure*}

%% file: figs/figures/custom_gans_full.tex
\begin{figure*}[!htb]
    \setlength{\tabcolsep}{2pt}
    \centering
    \begin{tabular}{p{1em}ccccc}
    \centering
    \rotatebox{90}{\;\;\; Image Example} &
    \includegraphics[width=0.185\textwidth, frame]{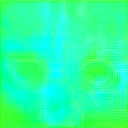}%
    & 
    \includegraphics[width=0.185\textwidth, frame]{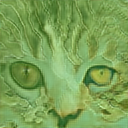}%
    & 
    \includegraphics[width=0.185\textwidth, frame]{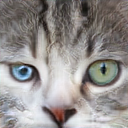}%
    & 
    \includegraphics[width=0.185\textwidth, frame]{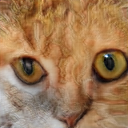}%
    & 
    \includegraphics[width=0.185\textwidth, frame]{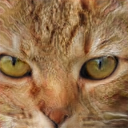} 
    \\%
    & Epoch 20 & Epoch 32 & Epoch 40 & Epoch 52 & Epoch 70
    \end{tabular}
    \caption{Examples of 
    images produced by $P_A$ ordered according to check point epochs. Starting from epoch 40 images retain high-quality.}
    \label{fig:catExamples}
    \centering
    \includegraphics[width=0.9\textwidth]{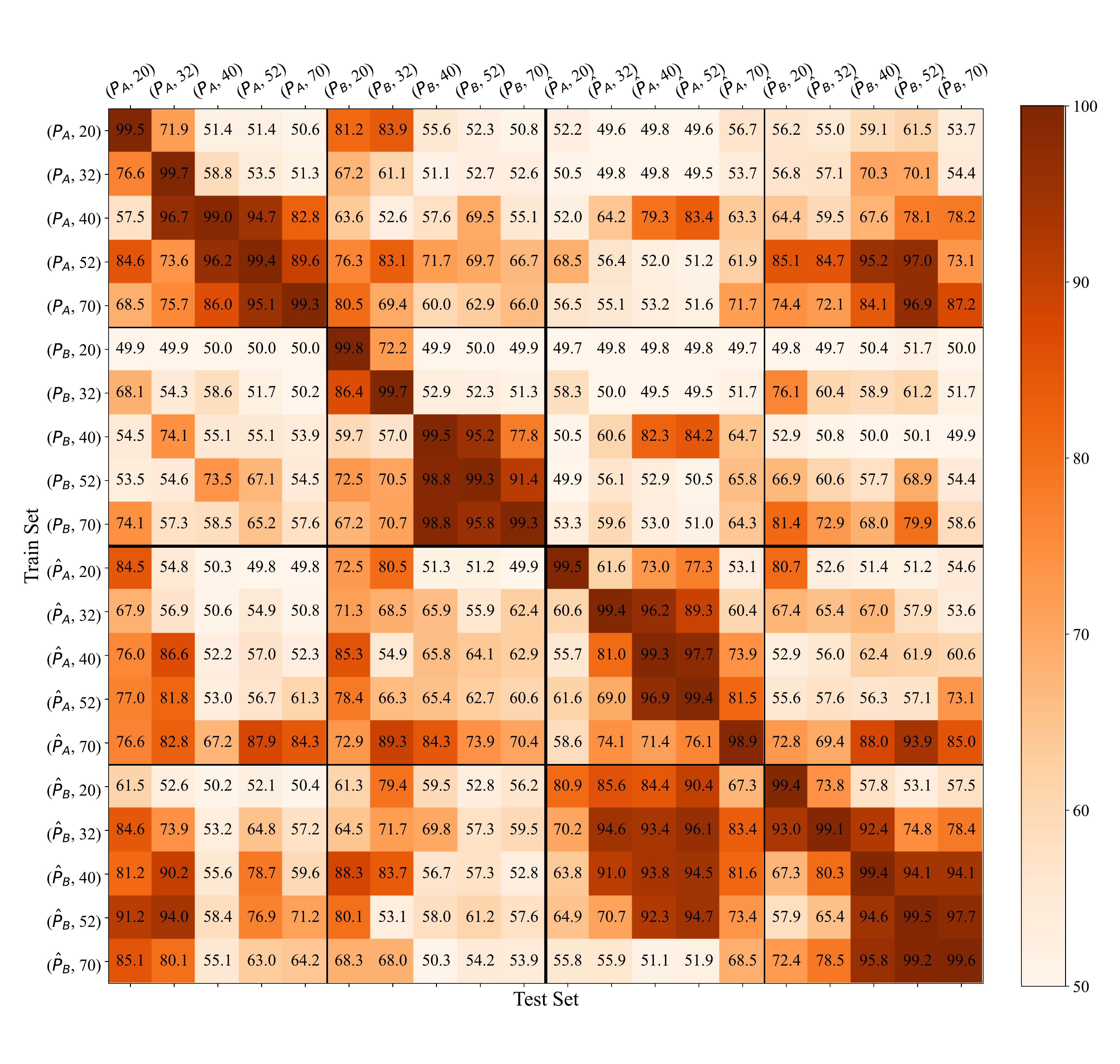}%
    \caption{Cross-detection accuracy (\%) for ProGAN models $P_A, P_B, \hat{P}_A$ and $\hat{P}_B$. Clusters of high cross-detection accuracy re-appear for each model at epoch 40,52 and 70.}
    \label{fig:crossevalCUSTOMF}
\end{figure*}

%% file: figs/figures/custom_gan_fingers.tex
\begin{figure*}[!htb]
     \centering
		\includegraphics[width=1.0\textwidth]{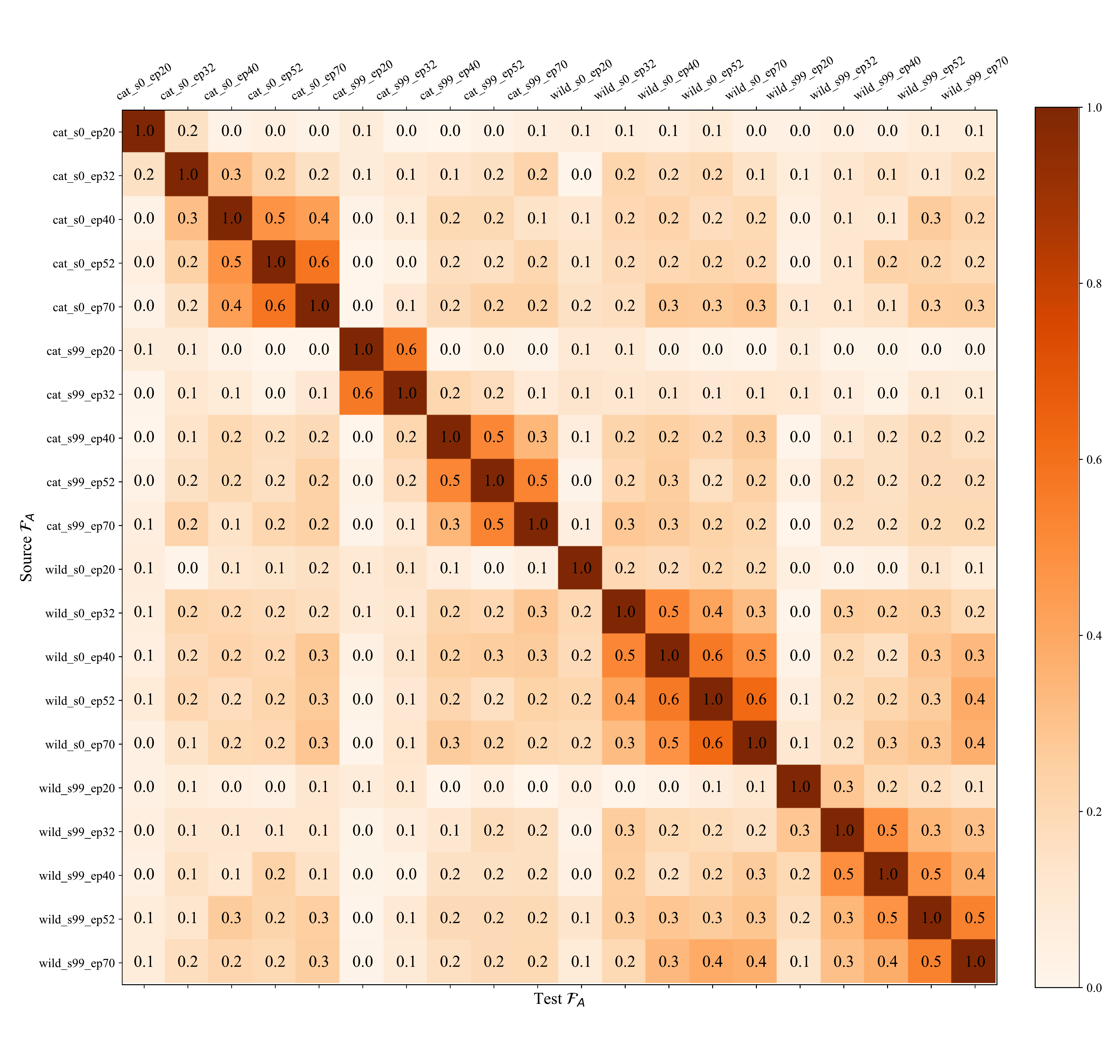}%
		\caption{Cross-correlation of $\mathcal{F}_A$ for ProGAN models $P_A, P_B, \hat{P}_A$ and $\hat{P}_B$. Values are symmetric for all the models and cluster of high values (above 0.5) are smaller then in \cref{fig:crossevalCUSTOMF}.}
		\label{fig:crossevalCUSTOMFA}
\end{figure*} 
\begin{figure*}[!htb]
     \centering
		\includegraphics[width=1.0\textwidth]{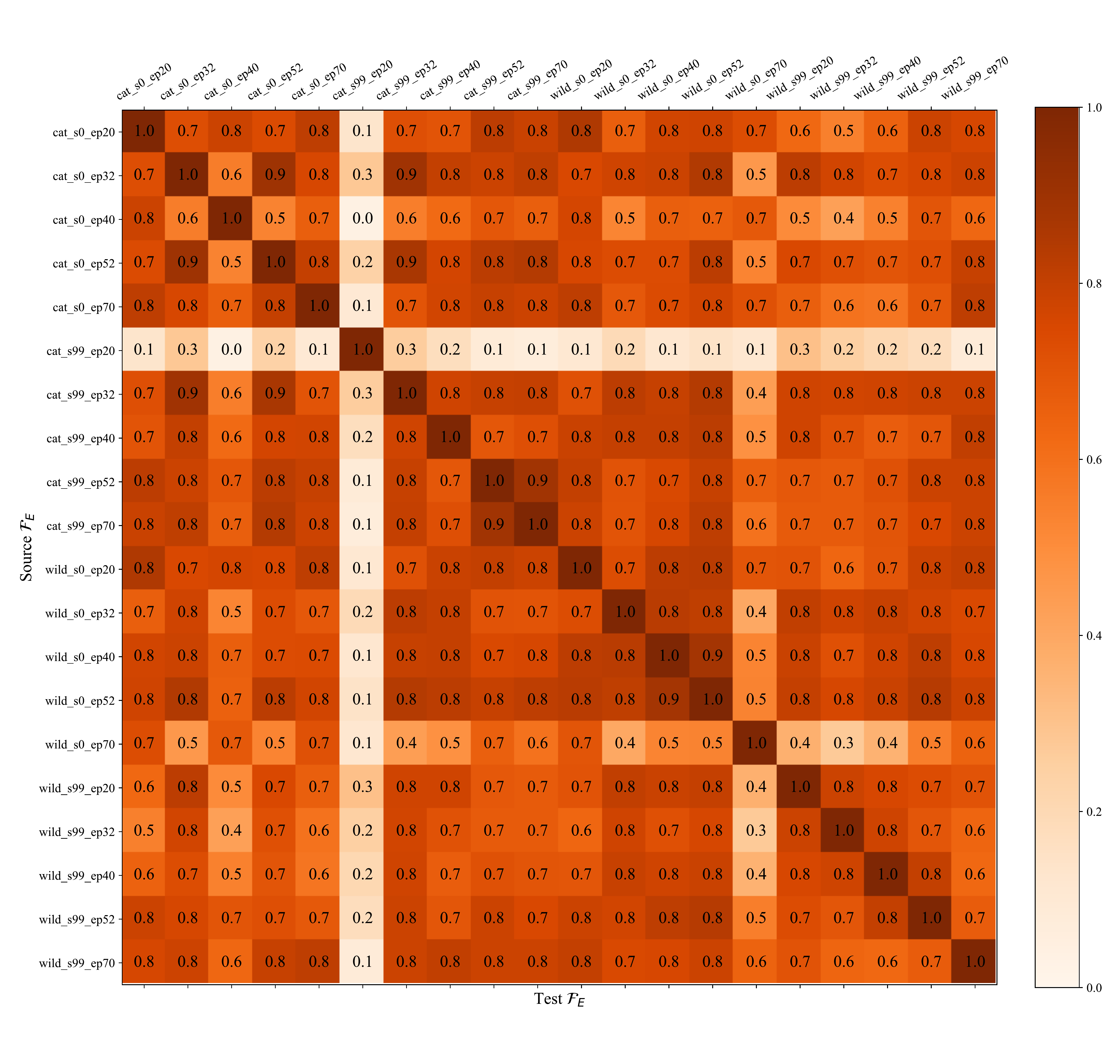}%
		\caption{Cross-correlation of $\mathcal{F}_E$ for ProGAN models $P_A, P_B, \hat{P}_A$ and $\hat{P}_B$. Relation appears to be random.}
		\label{fig:crossevalCUSTOMFE}
\end{figure*}